\documentclass[letterpaper, 10 pt, conference]{ieeeconf}  

\IEEEoverridecommandlockouts                              

\usepackage{amsmath}
\usepackage{amssymb}
\usepackage{amsfonts}
\usepackage{graphicx}
\usepackage{booktabs}
\usepackage{multirow}
\usepackage{cite}
\usepackage{url}

\title{\LARGE \bf
Image-to-Point Cloud Registration Made Easy\\with Rectified Flow-based LiDAR Upsampling
}

\author{Reon Tabata$^{1}$, Kenji Koide$^{2}$, Shuji Oishi$^{2}$, Masashi Yokozuka$^{2}$, \\Taku Okawara$^{2}$, Aoki Takanose$^{2}$, and Jun Miura$^{1}$
\thanks{$^{1}$Reon Tabata and Jun Miura are with Department of Computer Science and Engineering, Toyohashi University of Technology, Toyohashi, Aichi, Japan}
\thanks{$^{2}$Kenji Koide, Shuji Oishi, Masashi Yokozuka, Aoki Takanose, and Taku Okawara are with National Institute of Advanced Industrial Science and Technology, Tsukuba, Ibaraki, Japan}
}

\begin{document}
~\\
© 2026 IEEE.  Personal use of this material is permitted.  Permission from IEEE must be obtained for all other uses, in any current or future media, including reprinting/republishing this material for advertising or promotional purposes, creating new collective works, for resale or redistribution to servers or lists, or reuse of any copyrighted component of this work in other works.
\clearpage

\maketitle
\thispagestyle{empty}
\pagestyle{empty}

\begin{abstract}
Image-to-Point Cloud Registration (I2P) is essential for integrating camera and LiDAR in perception and autonomous systems, yet the modality gap between images and point clouds makes it difficult to achieve both high accuracy and strong generalization.
In this paper, we propose a simple yet effective I2P method that treats LiDAR as an imaging sensor: from a single sparse LiDAR scan, we generate a dense LiDAR intensity image using Conditional Rectified Flow, match it with a camera image using a pre-trained feature matcher, and estimate the 6-DoF relative pose via PnP-RANSAC.
The proposed model is pre-trained through a self-supervised image completion task and fine-tuned on a small amount of LiDAR data (neither image-point cloud pairs nor ground-truth sensor poses are required), enabling it to scale to diverse LiDAR and camera configurations.
Experiments on the R3LIVE dataset show that the proposed method achieves a mean error of 4.89° / 1.63\,m, outperforming existing methods, while completing a single registration in approximately 0.68\,s.
The project page is available at \url{https://smrg-students.github.io/iros2026_tabata_project_page/}.
\end{abstract}

\section{Introduction}

Image-to-Point Cloud Registration (I2P) is a key technique in autonomous systems and robotics for integrating camera and LiDAR modalities.
I2P enables many applications; for example, monocular-camera localization within a point cloud map allows easy and low-cost system deployment applicable to autonomous driving and robot navigation.
These applications require both high accuracy and strong generalization across diverse sensors and environments.

The typical I2P pipeline consists of 2D-3D correspondence estimation followed by pose estimation.
Most learning-based I2P methods require paired image-point cloud data with ground-truth poses, leading to high training cost and limited generalization\cite{feng20192d3d,ren2022corri2p,kang2024cofii2p,li20232d3d,li2021deepi2p,lv2021lccnet}.
Methods based on large-scale pre-trained models avoid this requirement but sacrifice accuracy and speed\cite{wang2023freereg}.
Therefore, the modality gap between images and point clouds makes it difficult to achieve both high accuracy and strong generalization across different LiDAR and camera models.

\begin{figure}[t]
  \centering
  \includegraphics[width=0.86\linewidth]{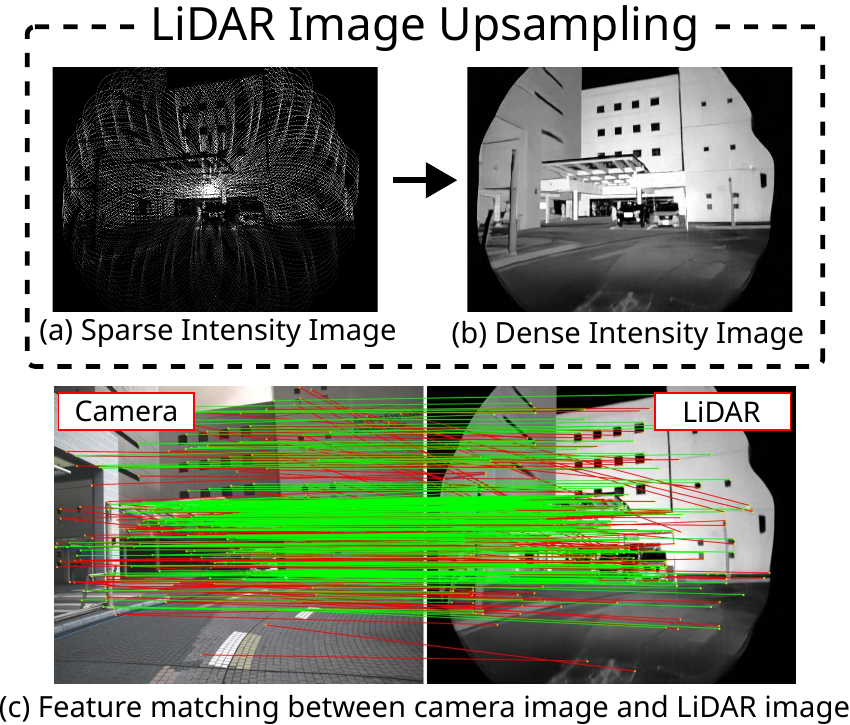}
  \caption{Proposed LiDAR upsampling and LiDAR-camera matching approach. (a) Sparse LiDAR intensity image created from a single LiDAR scan. (b) High-fidelity dense intensity image generated from the sparse scan by using Conditional Rectified Flow. (c) Feature correspondences between the upsampled LiDAR intensity image and the camera image.}
  \label{fig:teaser}
\end{figure}

In contrast, Image-to-Image Registration (I2I) has achieved notable performance recently.
Learning-based feature matching methods work under challenging conditions such as large viewpoint changes, large photometric gaps, and cross-modal (e.g., RGB vs thermal image matching)\cite{sarlin2020superglue, sun2021loftr, lindenberger2023lightglue, tuzcuouglu2024xoftr}.
In I2I, even when modalities differ, they can focus on local geometry pattern information and enable cross-domain learning across diverse environments and conditions, which helps achieve both high accuracy and good generalization.

Based on these observations, we propose a simple yet effective I2P method that treats LiDAR as an imaging sensor; from a single sparse point cloud, we generate a high-fidelity LiDAR intensity image and match it with a camera image (see Fig.~\ref{fig:teaser}).
This image-based formulation eliminates the need to learn explicit image-point cloud correspondences and allows us to leverage well-established pre-trained image feature matchers.
We project a LiDAR point cloud into a sparse intensity image, upsample it to a dense image, match it with the camera image, and estimate the 6-DoF relative pose.
For upsampling, we build on Rectified Flow\cite{liu2023flow}, which enables fast inference with fewer sampling steps than diffusion models (e.g., DDPM\cite{ho2020denoising}, DDIM\cite{song2020denoising}).
We use its variant, Conditional Rectified Flow, conditioned on the sparse intensity image.
Conditional Rectified Flow is pre-trained on image datasets with online masking to acquire inpainting capability in a self-supervised fashion.
Experiments demonstrate strong pose estimation and generalization performance without explicit paired image-point cloud supervision.

The main contributions of this work are three-fold:
\begin{itemize}
\item We propose a simple yet effective I2P registration method based on LiDAR image generation. By leveraging pre-trained image feature matchers, our approach avoids learning explicit image-point cloud matching, achieving high accuracy and strong generalization without paired image-point cloud training data that are expensive to collect.
\item We propose LiDAR image upsampling using Conditional Rectified Flow, which requires fewer sampling steps than other generative model-based methods. Even one-step generation shows little degradation and can run in about 0.68\,s. Thanks to the proposed pre-training procedure based on image completion, the model can generalize to different LiDAR models using only a small amount of LiDAR data.
\item On the R3LIVE dataset, although we did not use this dataset for training, our method achieved a mean error of 4.89° / 1.63\,m and outperformed baseline methods.
\end{itemize}

\section{Related Work}
\subsection{Image-to-Image Registration}
I2I has become a core technology for many tasks, such as localization and scene recognition.
In I2I, feature matching to find correspondences between images is a key component that largely determines overall performance.
Classical hand-crafted features (e.g., SIFT\cite{lowe2004distinctive}, SURF\cite{bay2006surf}, ORB\cite{rublee2011orb}) are still widely used for several tasks like Visual-SLAM (e.g., ORB-SLAM\cite{mur2015orb}).
Deep learning-based methods outperform classical approaches by learning visual priors from large datasets, eliminating the need for manual feature engineering.
SuperGlue\cite{sarlin2020superglue} formulates feature matching as an optimal transport problem solved by a graph neural network with cross-attention, and it remains robust under challenging conditions such as large viewpoint changes and large photometric gaps.
Several variants and faster feature matching methods have also been proposed (e.g., LoFTR\cite{sun2021loftr} and LightGlue\cite{lindenberger2023lightglue}); among them, some approaches explicitly adapt to cross-modal settings (e.g., thermal-to-RGB image matching), such as XoFTR\cite{tuzcuouglu2024xoftr}.
Overall, these learning-based approaches are highly effective, and it has also been reported that they can match RGB images to dense LiDAR intensity images\cite{koide2023general}.

Recently, correspondence-free approaches that simultaneously estimate camera poses and scene geometry in each image view as 3D point maps have emerged, such as DUSt3R\cite{wang2024dust3r}, VGGT\cite{wang2025vggt}, and MapAnything\cite{keetha2025mapanything}.
These approaches lessen the need for explicit correspondences and downstream geometric optimization, and can improve robustness to repetitive or low-texture scenes that challenge local feature matching.

Our current implementation employs the well-established feature matching model (LightGlue\cite{lindenberger2023lightglue}) because it exhibits a good cross-modal generalization.
It is worth noting that our proposed method is based on a general image generation pipeline, making it easy to extend to other feature matching methods or correspondence-free models with minor modifications and/or additional fine-tuning.

\begin{figure*}[t]
  \vspace{1.45mm}
  \centering
  \includegraphics[width=0.95\textwidth]{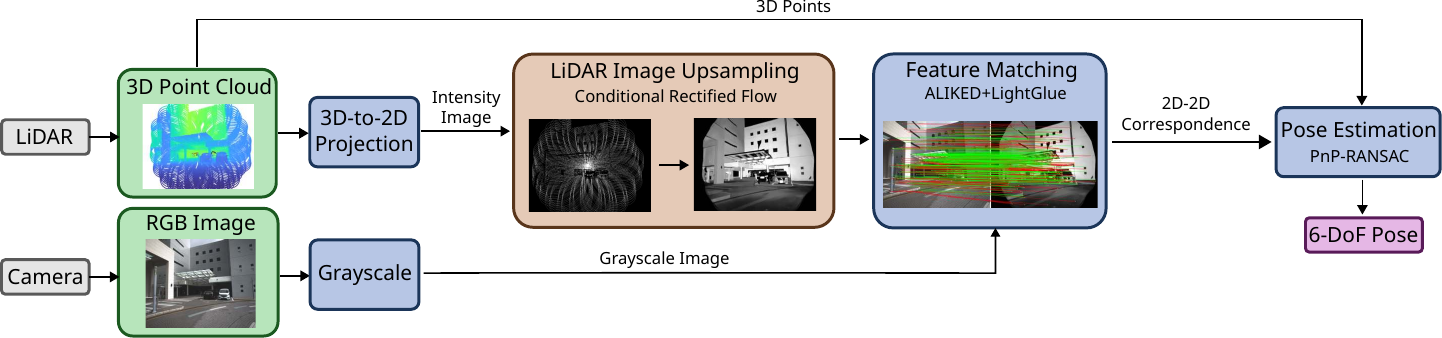}
  \caption{Overview of the proposed Image-to-Point Cloud registration pipeline. Given a sparse LiDAR scan and a camera image, the Conditional Rectified Flow module generates a dense LiDAR intensity image, which is then matched to the camera image via feature matching. The resulting 2D--3D correspondences are fed into PnP-RANSAC to estimate the 6-DoF relative pose.}
  \label{fig:method_overview}
\end{figure*}

\subsection{Image-to-Point Cloud Registration}
Despite the recent success of I2I, I2P is still challenging: LiDAR measurements are sparse with irregular scan patterns, making it difficult to learn general representations, especially when LiDAR models differ from training data or image-point cloud overlap is small.
Moreover, the modality gap between images and point clouds is large: although learning-based methods can reduce this gap, they often sacrifice generalization.
Existing I2P approaches can be broadly categorized into three groups: (i) methods that predict 2D-3D correspondences and estimate the pose via PnP, (ii) methods that avoid explicit 2D-3D correspondence construction, and (iii) methods that reduce the modality gap through large-scale pre-training.

Correspondence-based methods first establish 2D-3D correspondences between pixels and points, then estimate pose via PnP-RANSAC.
Early approaches such as 2D3D-MatchNet\cite{feng20192d3d} often suffered from low inlier ratios because modality differences make it difficult to reliably align keypoints.
To address this issue, CorrI2P\cite{ren2022corri2p} detects the overlapping region between the image and the point cloud and builds dense 2D-3D correspondences within that region.
CoFiI2P\cite{kang2024cofii2p} improves robustness by mitigating local minima in one-shot matching: it progressively refines correspondences from super-points/super-pixels to point/pixel-level matches.
In addition, 2D3D-MATR\cite{li20232d3d} proposes a coarse-to-fine framework that first matches image patches with point cloud patches and then densifies the correspondences.

Correspondence-free methods directly estimate pose from network predictions without establishing explicit 2D-3D correspondences.
DeepI2P\cite{li2021deepi2p} classifies whether each point lies inside the camera frustum and updates the pose based on these predictions, but it is often sensitive to classification errors near the field-of-view boundary.
LCCNet\cite{lv2021lccnet}, originally proposed for camera-LiDAR calibration, directly regresses the relative transform from a projected depth image and a camera image; however, it typically handles only a limited miscalibration range and is therefore less suitable for I2P settings with large viewpoint changes.
However, all these methods require paired image-point cloud datasets with global poses, and when trained on data with limited diversity in LiDAR scan patterns or viewpoints, they fail to generalize.

FreeReg\cite{wang2023freereg} uses large pre-trained models to bridge the modality gap without fine-tuning.
However, it requires many diffusion steps, leading to high latency (9.3\,s as reported in the original paper; 22.62\,s in our evaluation environment), making real-time use difficult.
Moreover, 6-DoF accuracy is limited by reliance on monocular depth estimation.


Motivated by the recent success of image-to-image matching models, we propose a simple yet effective I2P framework based on LiDAR image generation.
Since a single LiDAR scan produces a very sparse intensity image that is difficult to match directly against a camera image, we first upsample it into a dense image using a generative model, and then match the result to the camera image using a powerful pre-trained feature matcher.

\section{Methodology}

\subsection{Proposed Pipeline}
Our goal is to estimate the 6-DoF transformation ${}^C\boldsymbol{T}_L$ between the LiDAR and camera coordinate frames from a single pairing of a LiDAR point cloud and a camera image.
Let $\mathcal{P} = \{{}^L\boldsymbol{p}_i\}_{i=1}^N \subset \mathbb{R}^3$ be a single LiDAR scan with associated point intensities $\mathcal{L} = \{l_i\}_{i=1}^N \subset \mathbb{R}$, and let $\mathcal{I}(\boldsymbol{x}) = y$ be a camera image, where $\boldsymbol{x} \in \mathbb{R}^2$ are pixel coordinates and $y \in \mathbb{R}$ is the grayscale pixel value.
A point ${}^L\boldsymbol{p}_i$ in the LiDAR frame is transformed into the camera coordinate frame as ${}^C\boldsymbol{p}_i = {}^C\boldsymbol{T}_L \, {}^L\boldsymbol{p}_i$ and projected into the image plane via a projection function $\pi$: $\boldsymbol{x}_i = \pi({}^C\boldsymbol{p}_i)$.
To prioritize responsiveness and frame-by-frame independence, we avoid accumulation of LiDAR scans and multi-frame inputs, and estimation is performed on each single pairing $(\mathcal{P}, \mathcal{I})$.

The proposed pipeline proceeds in three stages, as illustrated in Fig.~\ref{fig:method_overview}.
First, $\mathcal{P}$ is projected into a sparse LiDAR intensity image $\mathcal{I}_L$ with a virtual camera model, which is then upsampled to a dense intensity image $\hat{\mathcal{I}}_L$ by a Conditional Rectified Flow model.
Second, a feature matcher extracts keypoints $\{\boldsymbol{u}_k\}_{k=1}^K \subset \mathbb{R}^2$ on the dense LiDAR intensity image $\hat{\mathcal{I}}_L$ and $\{\boldsymbol{v}_k\}_{k=1}^K \subset \mathbb{R}^2$ on the camera image $\mathcal{I}$, producing 2D--2D correspondences $\mathcal{M} = \{(\boldsymbol{u}_k, \boldsymbol{v}_k)\}_{k=1}^K$.
Finally, each LiDAR keypoint $\boldsymbol{u}_k$ is lifted to a 3D point ${}^L\boldsymbol{q}_k \in \mathbb{R}^3$ using the LiDAR depth, forming 2D--3D correspondences $\{(\boldsymbol{v}_k, {}^L\boldsymbol{q}_k)\}$ from which the LiDAR-camera transformation ${}^C\boldsymbol{T}_L$ is estimated via PnP-RANSAC\cite{fischler1981random,lepetit2009ep}.

\subsection{LiDAR Data Upsampling}
The LiDAR upsampling module projects the point cloud $\mathcal{P}$ with intensities $\mathcal{L}$ into a sparse LiDAR intensity image $\mathcal{I}_L$, and then outputs a dense intensity image $\hat{\mathcal{I}}_L$.
Conditional diffusion models have been shown to be effective for this task~\cite{nakashima2024lidar}, where point clouds are projected to range and intensity images and restored via image inpainting.
Rectified Flow~\cite{liu2023flow}, which transports samples along straight-line trajectories and enables high-quality generation with fewer inference steps, has further been applied to LiDAR image generation~\cite{nakashima2025fast}.
Building on these advances, we adopt Conditional Rectified Flow as our upsampling model, which generalizes well with limited fine-tuning data to adapt to various LiDAR sensors while also keeping generation time small.

A sparse LiDAR intensity image $\mathcal{I}_L$ is projected from a single LiDAR scan $\mathcal{P}$ using a virtual camera model. 
While it is agnostic to the projection model, we use the pinhole model here; an equirectangular model can be used for omnidirectional cameras.
As shown in Fig.~\ref{fig:teaser}(a), a single LiDAR scan $\mathcal{P}$ is extremely sparse, making it difficult to directly extract meaningful information from $\mathcal{I}_L$.
By conditioning Conditional Rectified Flow on $\mathcal{I}_L$, we obtain the dense image $\hat{\mathcal{I}}_L$ from any single scan.
To prioritize model generality and geometric correctness during matching, we intentionally avoid depth completion and perform completion only in appearance.
We found that the dense intensity image $\hat{\mathcal{I}}_L$ generated by Conditional Rectified Flow is not always accurate in fine details, but it preserves sufficient global consistency and is reliable enough for image matching.

Moreover, the proposed Conditional Rectified Flow model is pre-trained in a self-supervised fashion via image inpainting on RGB datasets, independent of any LiDAR-specific scan patterns.
Therefore, Conditional Rectified Flow can be adapted to various LiDAR sensors with only minimal fine-tuning.
Implementation details are described in Section \ref{sec:conditional_rectified_flow}.

\subsection{Feature Matching}
Given a dense intensity image $\hat{\mathcal{I}}_L$ and the camera image $\mathcal{I}$, we perform feature matching to obtain 2D--2D correspondences $\mathcal{M} = \{(\boldsymbol{u}_k, \boldsymbol{v}_k)\}_{k=1}^K$, where $\boldsymbol{u}_k \in \hat{\mathcal{I}}_L$ and $\boldsymbol{v}_k \in \mathcal{I}$ are keypoints in the LiDAR and camera images, respectively.
Since $\hat{\mathcal{I}}_L$ exhibits a similar appearance to a grayscale camera image, the proposed method matches $\hat{\mathcal{I}}_L$ to $\mathcal{I}$ using a cross-modal image feature matching model.
The proposed pipeline is agnostic to the matching algorithm, and any model can be used, for example SuperGlue\cite{sarlin2020superglue} or LightGlue\cite{lindenberger2023lightglue}.
To balance speed and accuracy, we use pre-trained ALIKED\cite{zhao2023aliked} for feature detection and LightGlue\cite{lindenberger2023lightglue} for matching, with no fine-tuning.

Since identifying correct cross-modal correspondences is generally difficult due to appearance differences, we set a low matching threshold to obtain many correspondences.
In this setting, we obtain a large number of correspondences $\mathcal{M}$, but also many false matches. However, these outliers can be effectively eliminated by the RANSAC-based pose estimation.

\subsection{PnP Pose Estimation}
Finally, we estimate the LiDAR-to-camera relative pose ${}^C\boldsymbol{T}_L$ from 2D--3D correspondences $\{(\boldsymbol{v}_k, {}^L\boldsymbol{q}_k)\}$.
Since cross-modal matching between $\mathcal{I}$ and $\hat{\mathcal{I}}_L$ may contain many outliers,
we use PnP-RANSAC\cite{fischler1981random,lepetit2009ep} to reduce the influence of wrong matches.
The correspondences are pairs of camera keypoints $\boldsymbol{v}_k \in \mathcal{I}$ and LiDAR keypoints $\boldsymbol{u}_k \in \hat{\mathcal{I}}_L$.
To apply PnP, we recover 3D points ${}^L\boldsymbol{q}_k$ (expressed in the LiDAR frame) for each LiDAR keypoint $\boldsymbol{u}_k$ using the LiDAR depth.
Since only intensity is upsampled (not depth), PnP-RANSAC is applied only to matches $(\boldsymbol{u}_k, \boldsymbol{v}_k)$ for which LiDAR depth exists in the local neighborhood of $\boldsymbol{u}_k$.
Since LiDAR data are sparse, using depth from distant pixels can cause large errors; unless otherwise stated, we use a 3-pixel radius and discard matches with no depth within this radius.

\section{Conditional Rectified Flow} \label{sec:conditional_rectified_flow}
Rectified Flow is a generative model that transports samples from a source distribution $p_{\mathrm{src}} = \mathcal{N}(\boldsymbol{0}, \boldsymbol{I})$ to a target distribution $p_{\mathrm{tgt}}$ by solving the ordinary differential equation (ODE) \cite{liu2023flow}:
\begin{equation}
  \frac{d\boldsymbol{X}_t}{dt} = \boldsymbol{V}_\theta(\boldsymbol{X}_t, t), \quad t \in [0, 1],
\end{equation}
where $\boldsymbol{X}_t \in \mathbb{R}^{H \times W}$ is the image state at time $t$, and $\boldsymbol{V}_\theta$ is a learned velocity field.
Unlike DDPMs, which model the reverse of a stochastic noising process $\boldsymbol{X}_t = \sqrt{\bar{\alpha}_t}\,\boldsymbol{X}_{\mathrm{tgt}} + \sqrt{1-\bar{\alpha}_t}\,\boldsymbol{\epsilon}$ ($\boldsymbol{\epsilon} \sim \mathcal{N}(\boldsymbol{0},\boldsymbol{I})$) and require a large number of stochastic sampling steps, Rectified Flow learns a deterministic flow along straight-line paths $\boldsymbol{X}_t = (1-t)\boldsymbol{X}_{\mathrm{src}} + t\,\boldsymbol{X}_{\mathrm{tgt}}$.
This straight-line transport enables high-quality generation with far fewer inference steps.
In this work, we extend Rectified Flow to conditional generation by conditioning $\boldsymbol{V}_\theta$ on the sparse LiDAR intensity image $\mathcal{I}_L$, yielding the dense image $\hat{\mathcal{I}}_L$ with low latency.

\subsection{Architecture}
\begin{figure}[t]
  \vspace{1.45mm}
  \centering
  \includegraphics[width=0.90\linewidth]{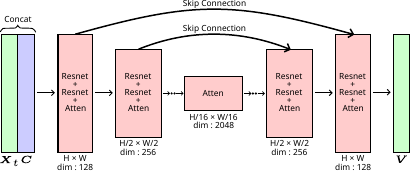}
  \caption{Conditional Rectified Flow architecture used for LiDAR intensity upsampling. U-Net backbone with added self-attention.}
  \label{fig:model_arch}
\end{figure}

We build a Conditional Rectified Flow model based on SR3\cite{saharia2022image}, a diffusion-based super-resolution method with a U-Net architecture\cite{ronneberger2015u}.
We add self-attention layers after ResNet blocks to improve global image understanding, as shown in Fig.~\ref{fig:model_arch}.
The model takes image $\boldsymbol{X}_t$ at time $t$ and condition image $\boldsymbol{C}$ as input, outputting velocity field $\boldsymbol{V}_\theta$.
Integrating along the ODE generates the target image.
We concatenate $\boldsymbol{X}_t$ and $\boldsymbol{C}$ following SR3.
We use single-scan LiDAR intensity $\mathcal{I}_L$ as $\boldsymbol{C}$ to generate dense image $\hat{\mathcal{I}}_L$.

\subsection{Loss Function and Sampling}
The loss function follows that of standard Rectified Flow \cite{liu2023flow}, assuming single-channel $H \times W$ images as input.
Let $\boldsymbol{X}_{\mathrm{src}} \in \mathbb{R}^{H \times W}$ be a Gaussian noise image and $\boldsymbol{X}_{\mathrm{tgt}} \in \mathbb{R}^{H \times W}$ be the target dense intensity image.
The interpolated image at time $t \in [0, 1]$ is defined as:
\[
\boldsymbol{X}_t = (1-t)\,\boldsymbol{X}_{\mathrm{src}} + t\,\boldsymbol{X}_{\mathrm{tgt}}, \quad \boldsymbol{X}_t \in \mathbb{R}^{H \times W}
\]
The model $\boldsymbol{V}_\theta(\boldsymbol{X}_t, \boldsymbol{C}, t) \in \mathbb{R}^{H \times W}$ predicts the velocity at time $t$,
conditioned on $\boldsymbol{C} \in \mathbb{R}^{H \times W}$, the sparse LiDAR intensity image.
The training loss is:
\[
L(\theta) = \mathbb{E}_{\boldsymbol{X}_{\mathrm{src}},\,\boldsymbol{X}_{\mathrm{tgt}},\,t}\!\left[\left\|\,\boldsymbol{X}_{\mathrm{tgt}} - \boldsymbol{X}_{\mathrm{src}} - \boldsymbol{V}_\theta(\boldsymbol{X}_t,\,\boldsymbol{C},\,t)\,\right\|_2^2\right]
\]
This trains the model to predict the true displacement $\boldsymbol{X}_{\mathrm{tgt}} - \boldsymbol{X}_{\mathrm{src}}$ at each interpolated state $\boldsymbol{X}_t$,
which approximates the velocity field that transports samples from $p(\boldsymbol{X}_{\mathrm{src}})$ to $p(\boldsymbol{X}_{\mathrm{tgt}})$.

During generation, the model integrates the velocity field from a Gaussian noise image $\boldsymbol{X}_{\mathrm{src}}$, discretized into $N$ Euler steps with step size $\Delta t = 1/N$:
\[
\boldsymbol{X}_{k+1} = \boldsymbol{X}_k + \boldsymbol{V}_\theta\!\bigl(\boldsymbol{X}_k,\,\boldsymbol{C},\,t_k\bigr)\,\Delta t, \quad t_k = k\,\Delta t
\]

\subsection{Training}
\subsubsection{Pre-Training}
\begin{figure}[t]
  \vspace{1.45mm}
  \centering
  \begin{tabular}{@{}ccc@{}}
    \includegraphics[width=0.32\linewidth]{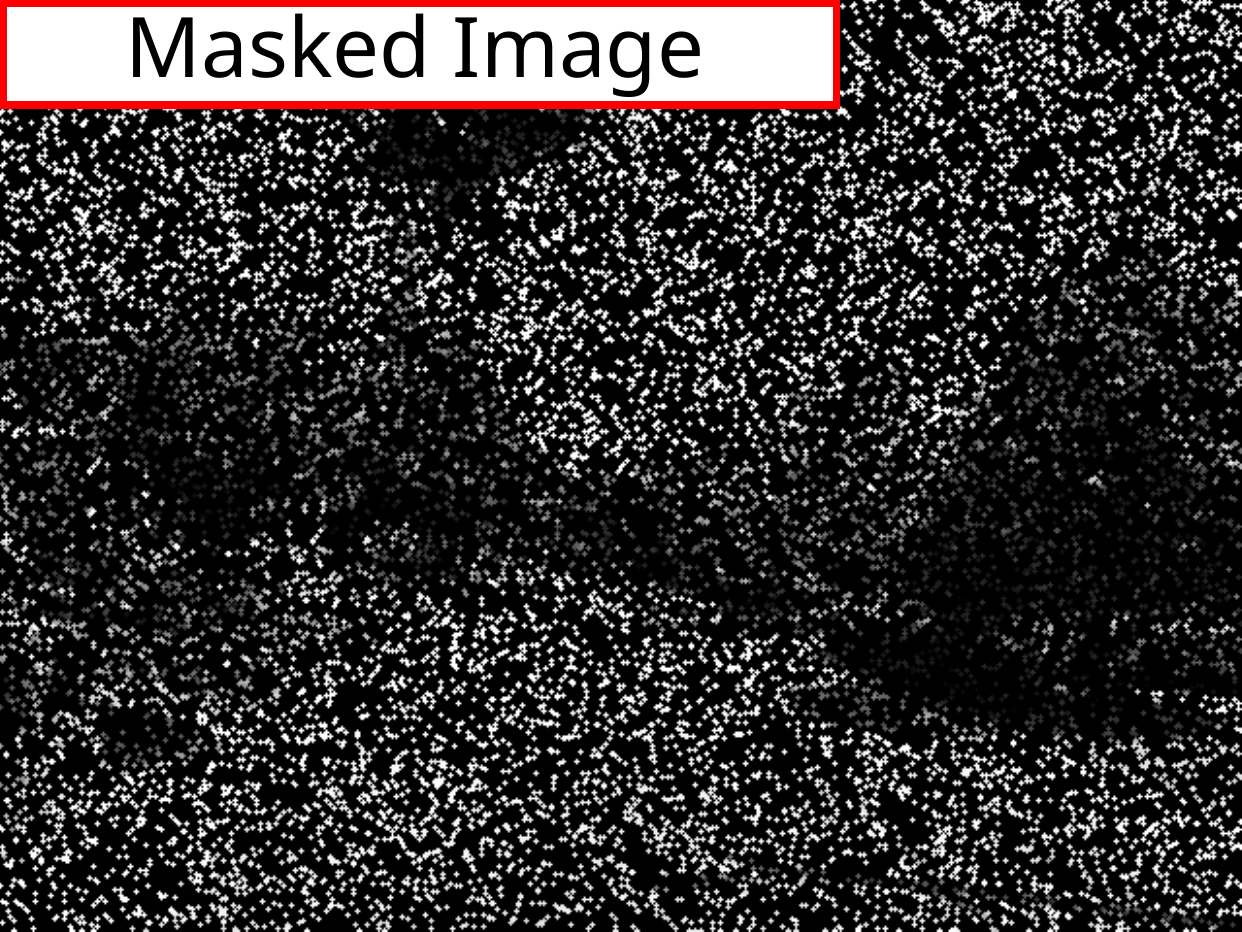} &
    \includegraphics[width=0.32\linewidth]{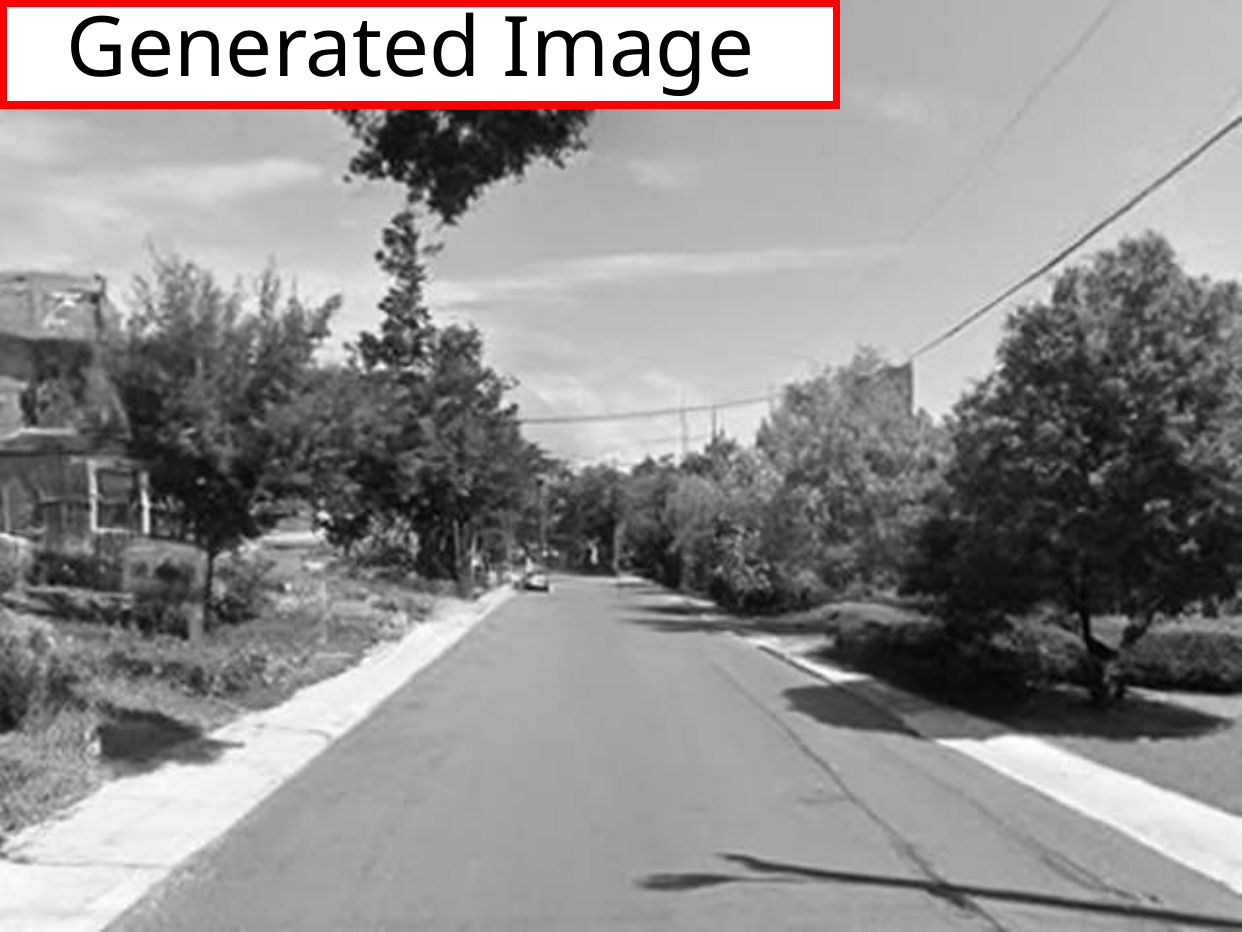} &
    \includegraphics[width=0.32\linewidth]{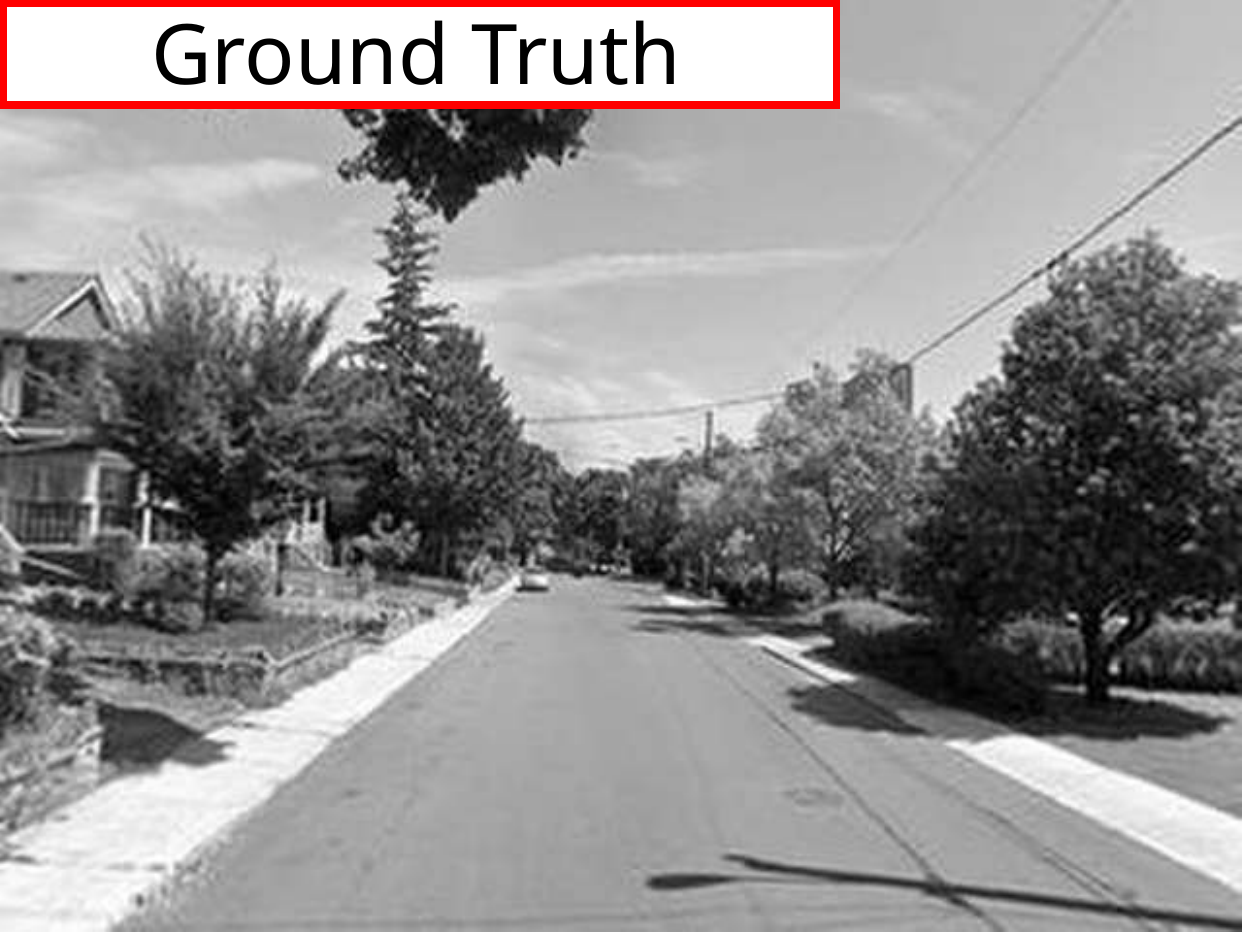} \\[-0.8ex]
    \multicolumn{3}{c}{(a) random point masking} \\[0.8ex]
    \includegraphics[width=0.32\linewidth]{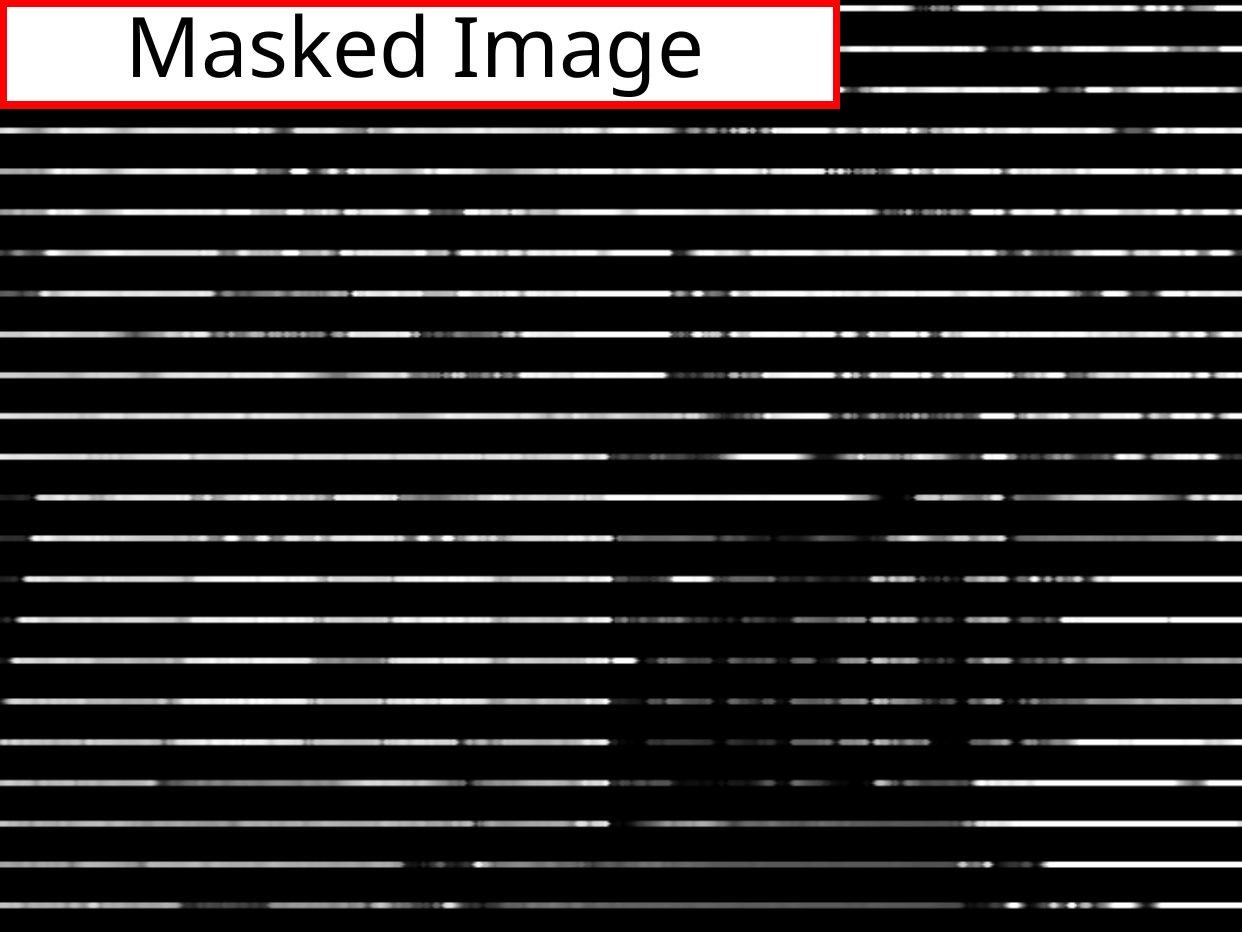} &
    \includegraphics[width=0.32\linewidth]{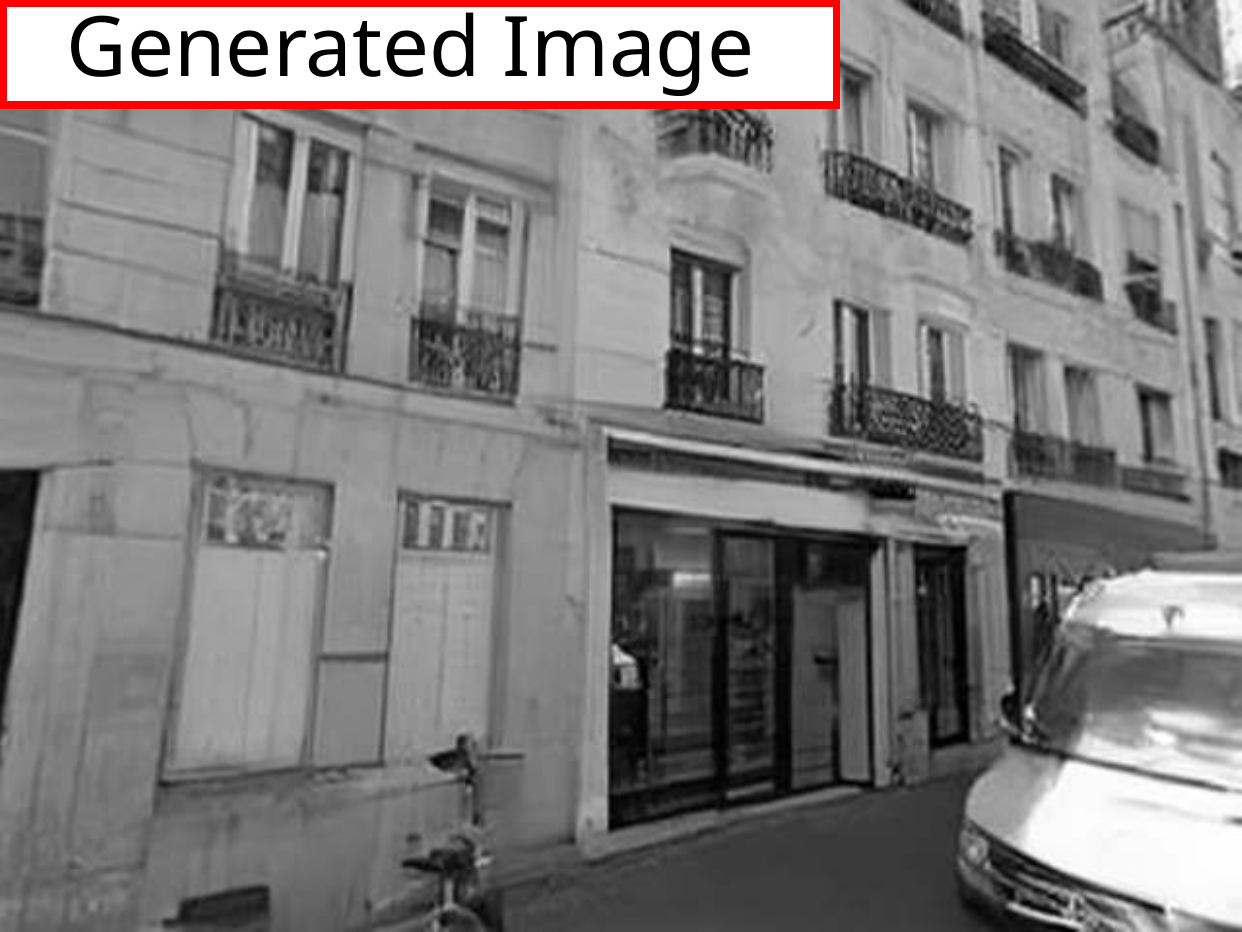} &
    \includegraphics[width=0.32\linewidth]{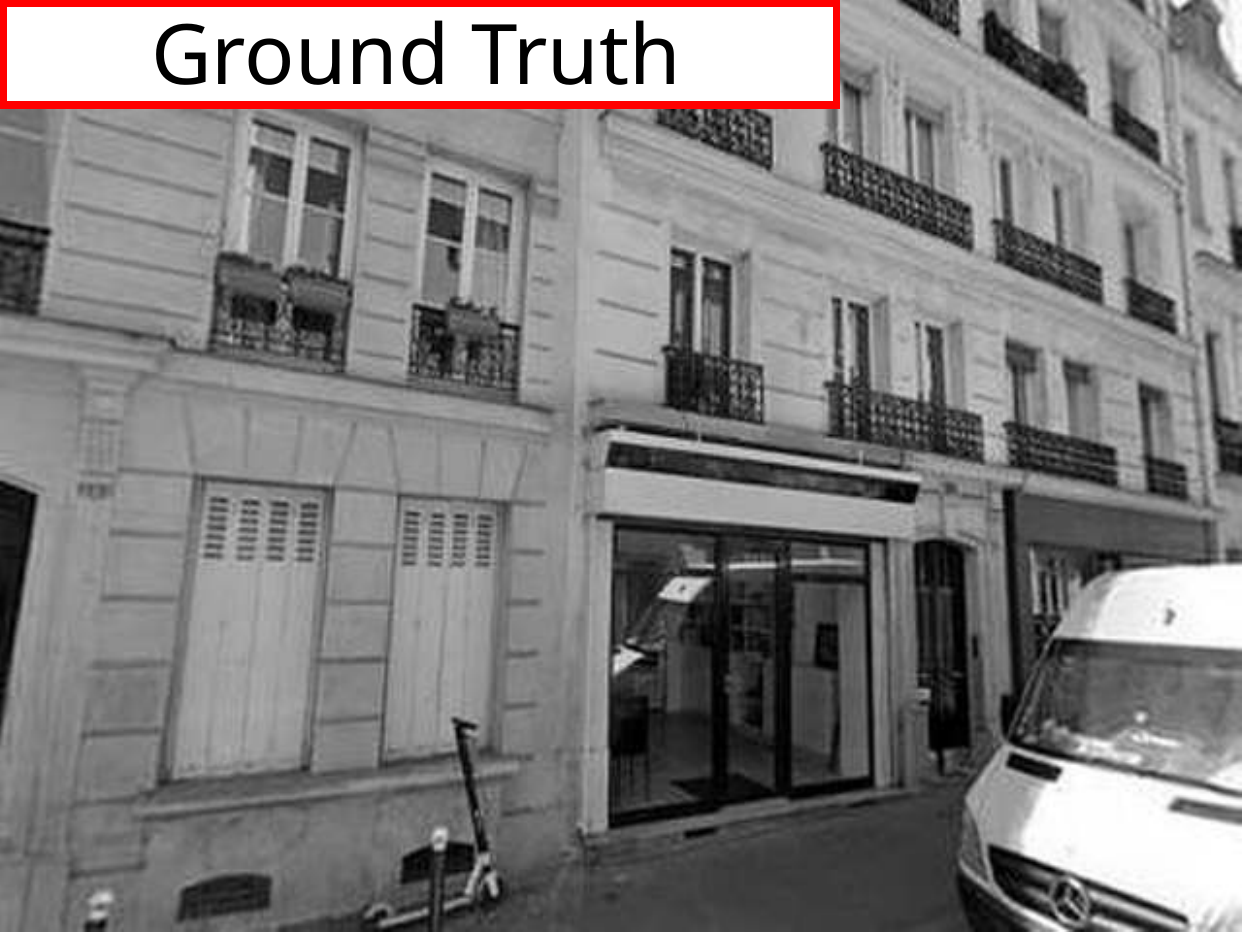} \\[-0.8ex]
    \multicolumn{3}{c}{(b) horizontal line masking} \\[0.8ex]
    \includegraphics[width=0.32\linewidth]{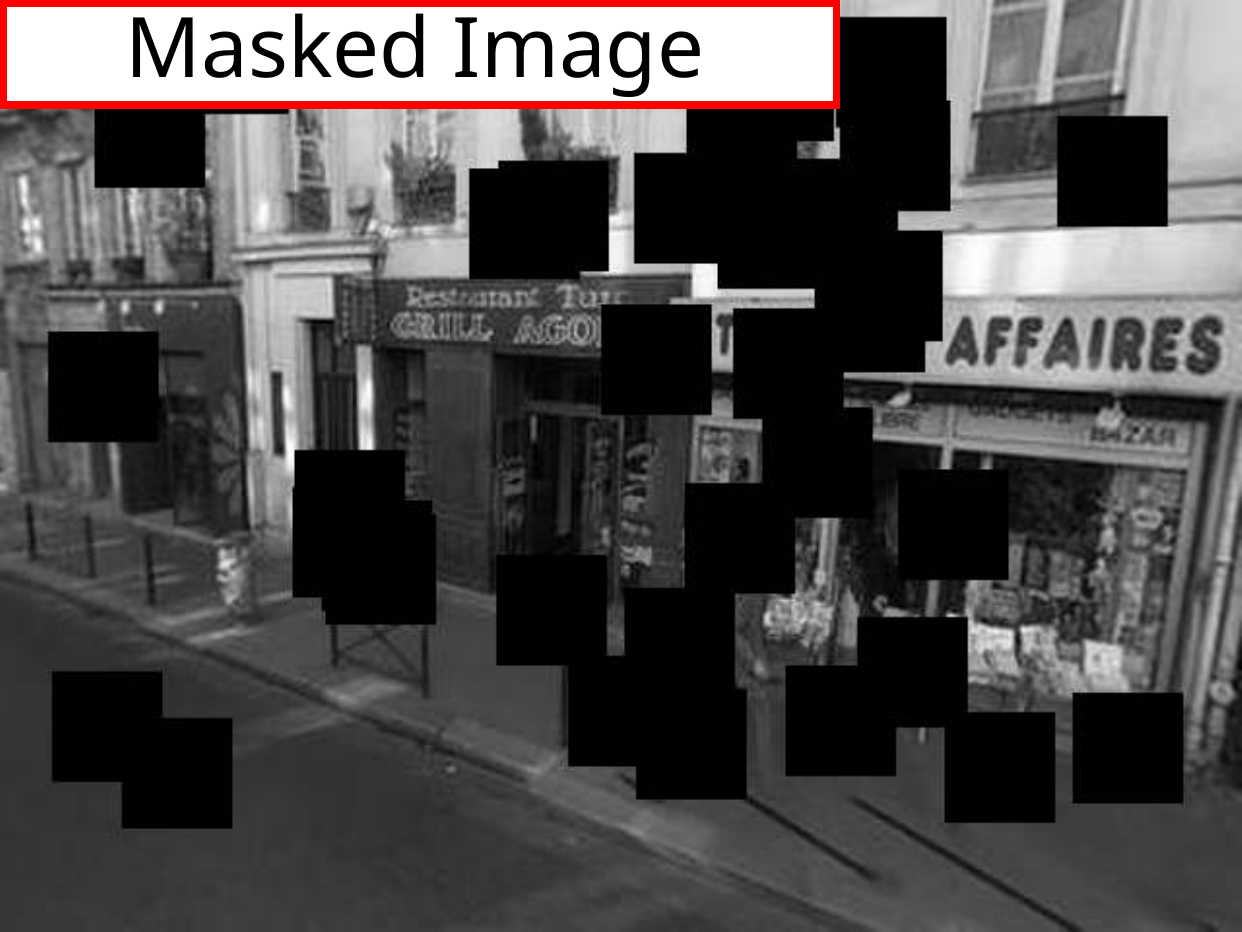} &
    \includegraphics[width=0.32\linewidth]{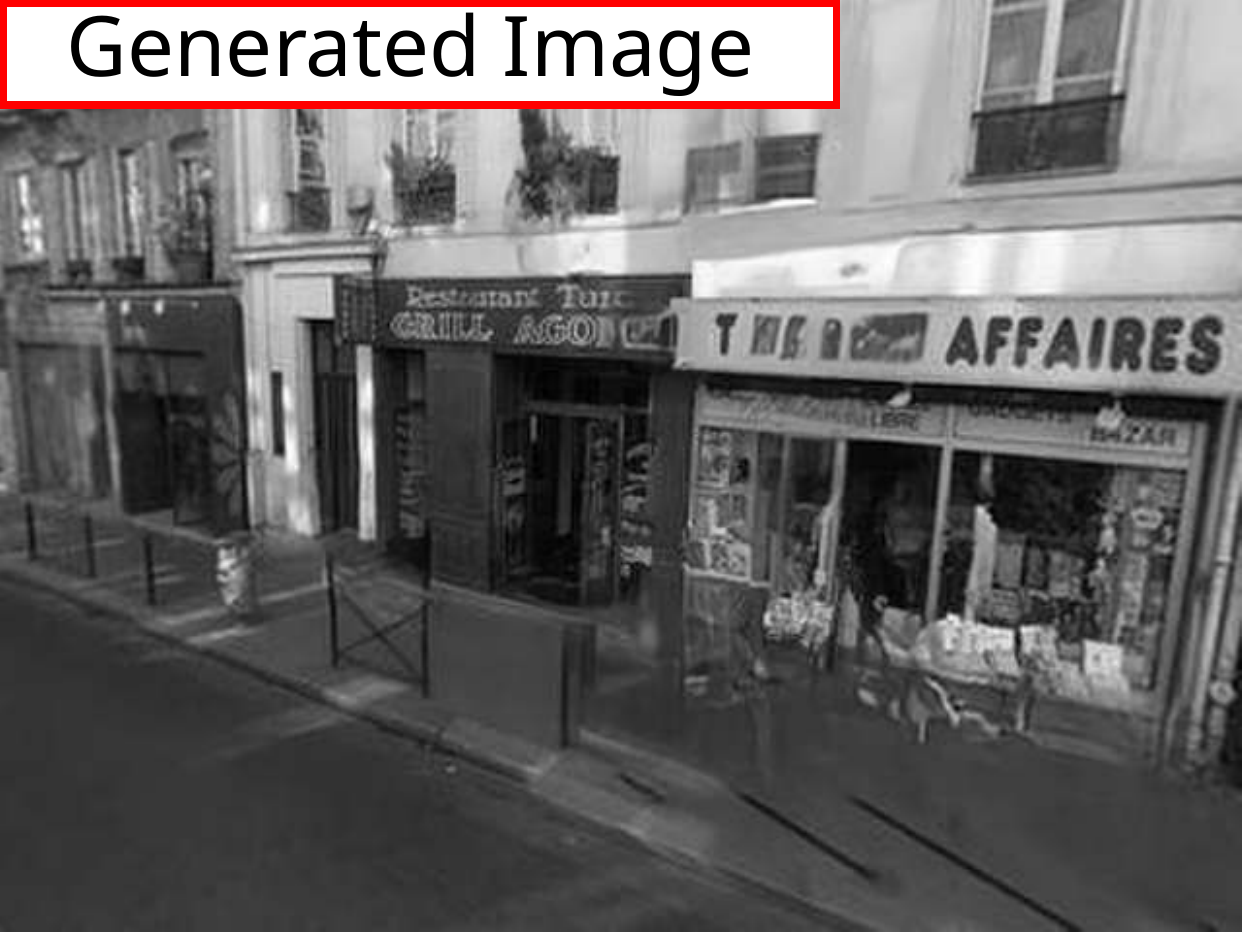} &
    \includegraphics[width=0.32\linewidth]{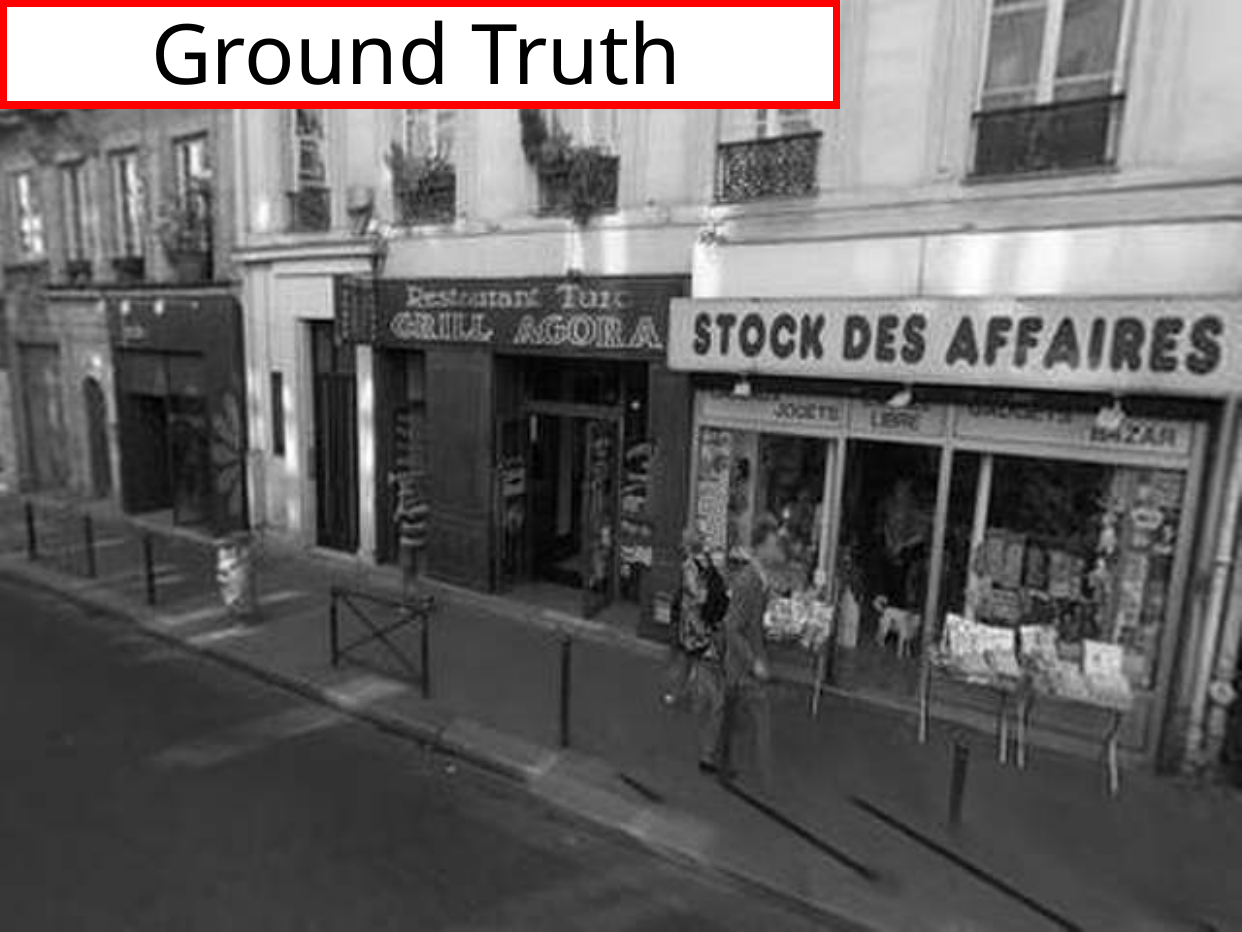} \\[-0.8ex]
    \multicolumn{3}{c}{(c) random block masking}
  \end{tabular}
  \caption{Pre-training examples with online masking. Each row shows a different masking pattern: (a) random point, (b) horizontal line, and (c) random block. Columns show (left) masked input, (middle) generated image, and (right) ground truth.}
  \label{fig:pre-train}
\end{figure}

Since dense LiDAR intensity images are difficult to collect at scale, we pre-train the model on image inpainting and fine-tune it for LiDAR intensity images.
The pre-training task restores a grayscale image from its masked version, teaching the model to complement sparse scan patterns.
We use the GSV-Cities\cite{ali2022gsv} dataset with four on-the-fly masking patterns: random point, vertical line, horizontal line, and block masking.
Random point masking (see Fig.~\ref{fig:pre-train}(a)) removes individual pixels, vertical and horizontal line masking (see Fig.~\ref{fig:pre-train}(b)) leave evenly-spaced lines, and block masking (see Fig.~\ref{fig:pre-train}(c)) place square patches at random positions.
The parameters of these masks (occlusion rate, line spacing, block count, and block size) are randomly sampled at each training iteration to increase pattern diversity.
The condition input of Conditional Rectified Flow is the masked image, and the reference is the original grayscale image.
To improve generalization across LiDAR models, we avoid LiDAR-specific scan patterns and use only generic masks.

Random and line masks require the ability to restore images locally using information from unmasked regions, while block masks require predicting the overall structure of masked areas.
As shown in Fig.~\ref{fig:pre-train}, fine details are restored for random and line masks (a)--(b), and the overall structure is reconstructed for block masks (c), confirming the model's ability to leverage surrounding context.

\subsubsection{Fine-Tuning}
\begin{figure}[t]
  \vspace{1.45mm}
  \centering
  \begin{tabular}{@{}ccc@{}}
    \includegraphics[width=0.32\linewidth]{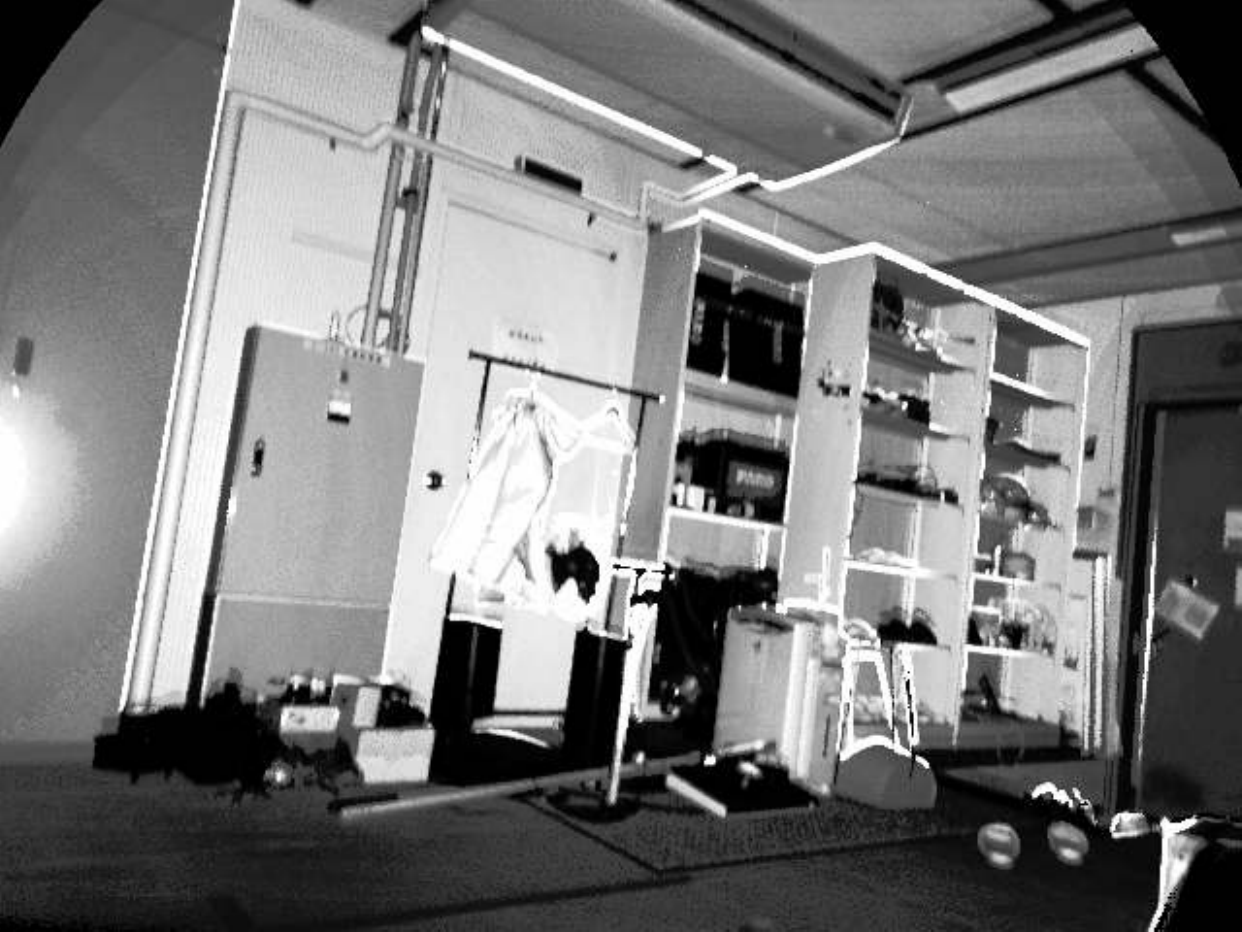} &
    \includegraphics[width=0.32\linewidth]{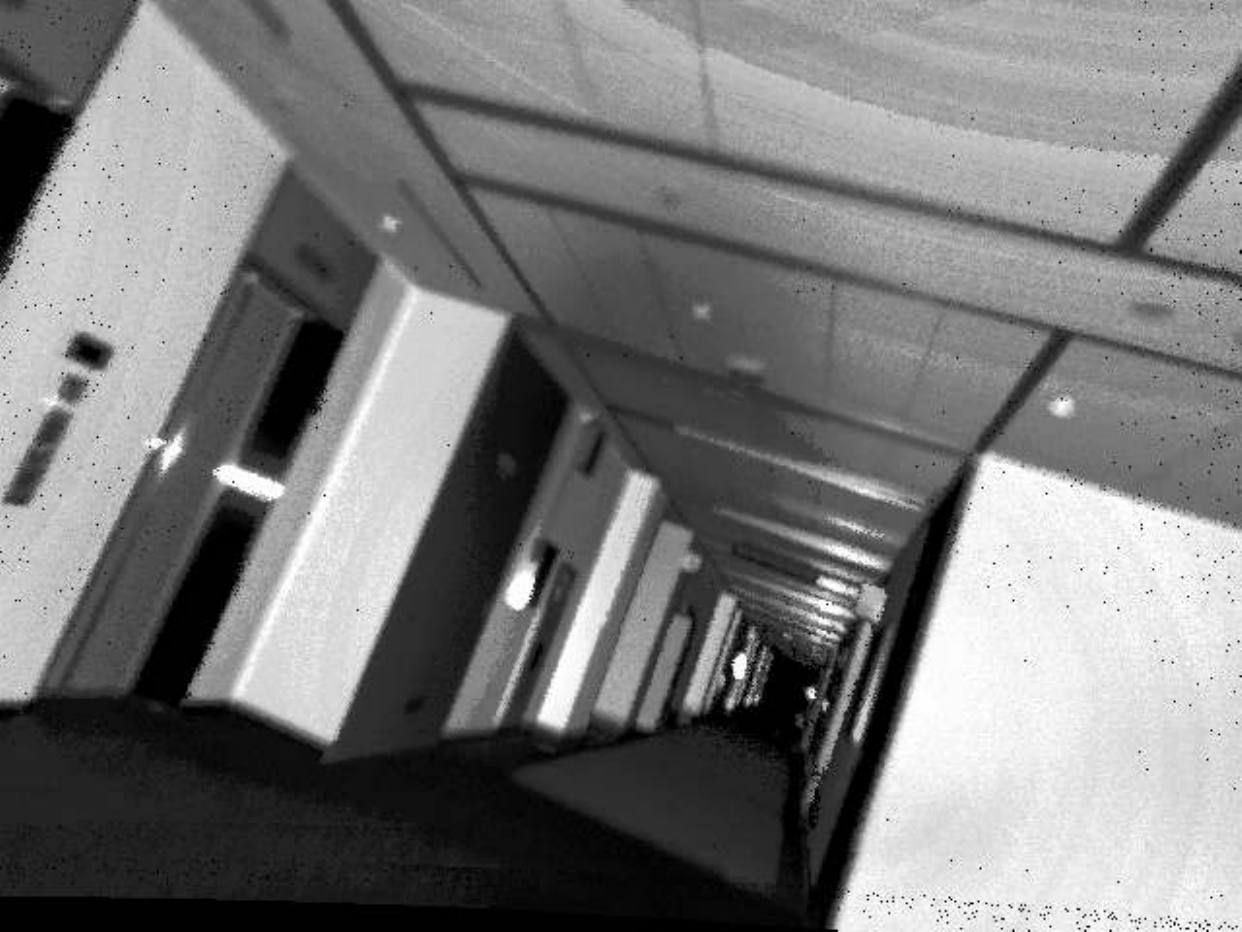} &
    \includegraphics[width=0.32\linewidth]{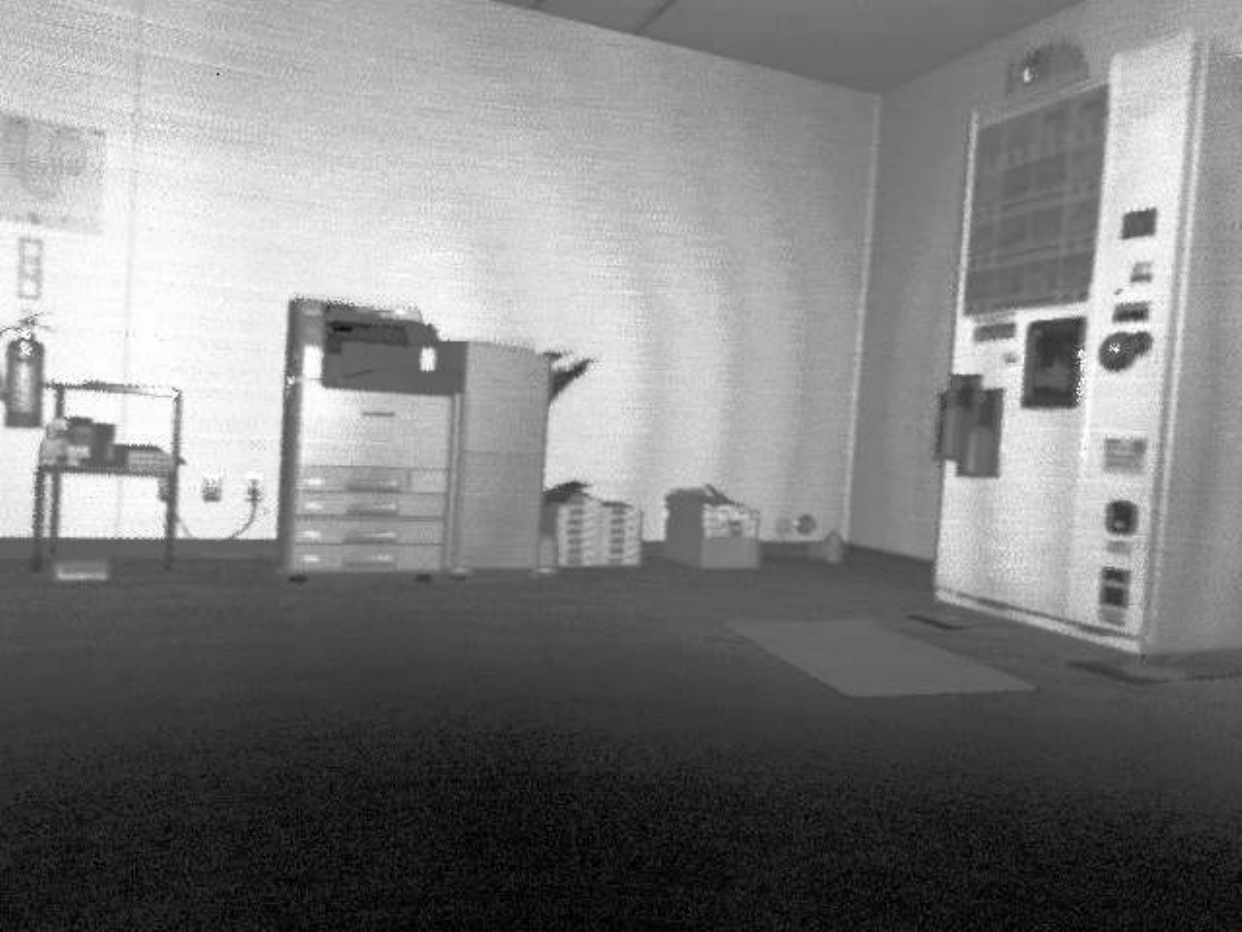} \\[-0.8ex]
    (a) AVIA & (b) MID360 & (c) OS1-64
  \end{tabular}
  \caption{Dense LiDAR intensity images collected for fine-tuning from three sensor types: (a) Livox AVIA, (b) Livox MID360, and (c) Ouster OS1-64.}
  \label{fig:intensity_images}
\end{figure}

\begin{figure}[t]
  \centering
  \begin{tabular}{@{}ccc@{}}
    \includegraphics[width=0.32\linewidth]{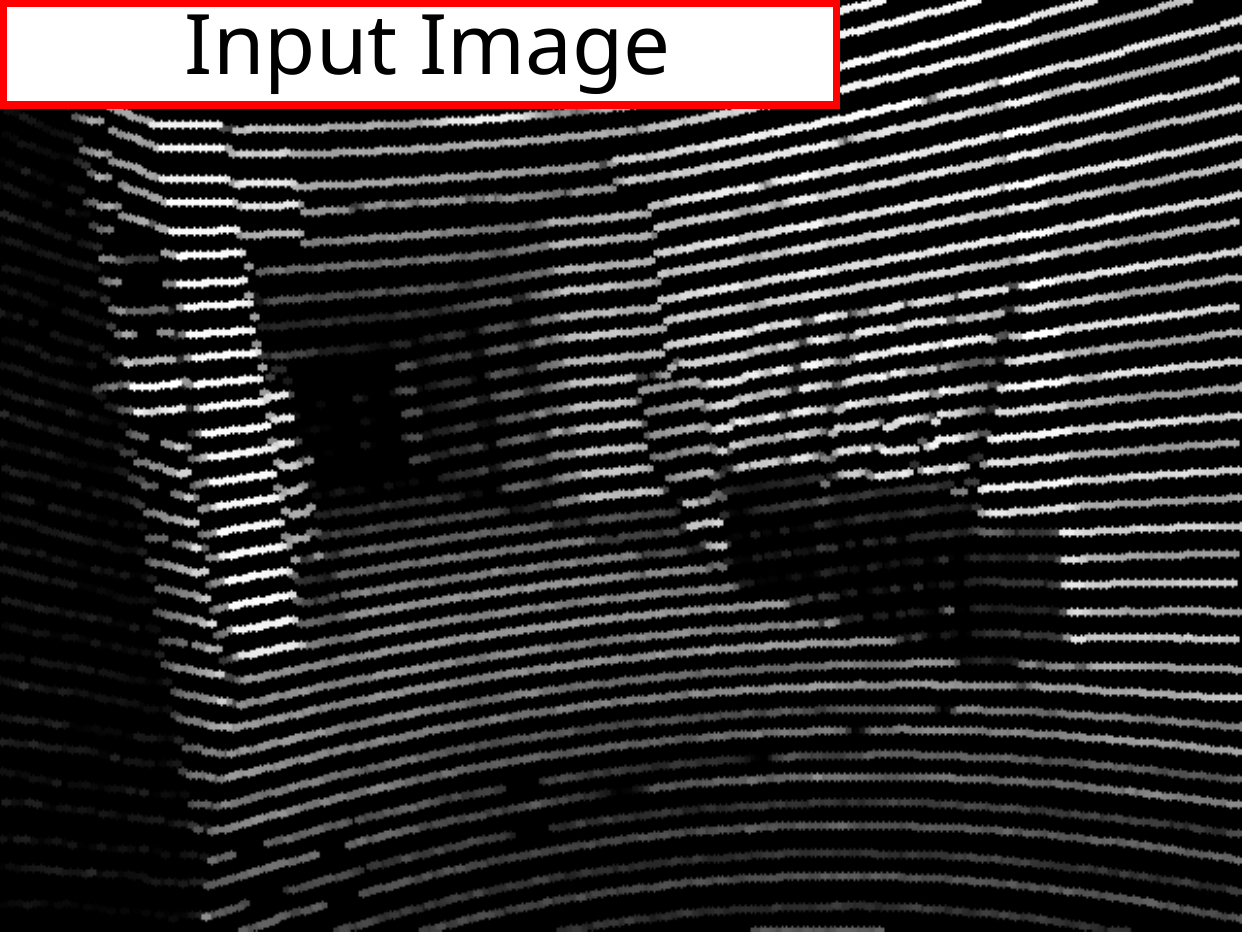} &
    \includegraphics[width=0.32\linewidth]{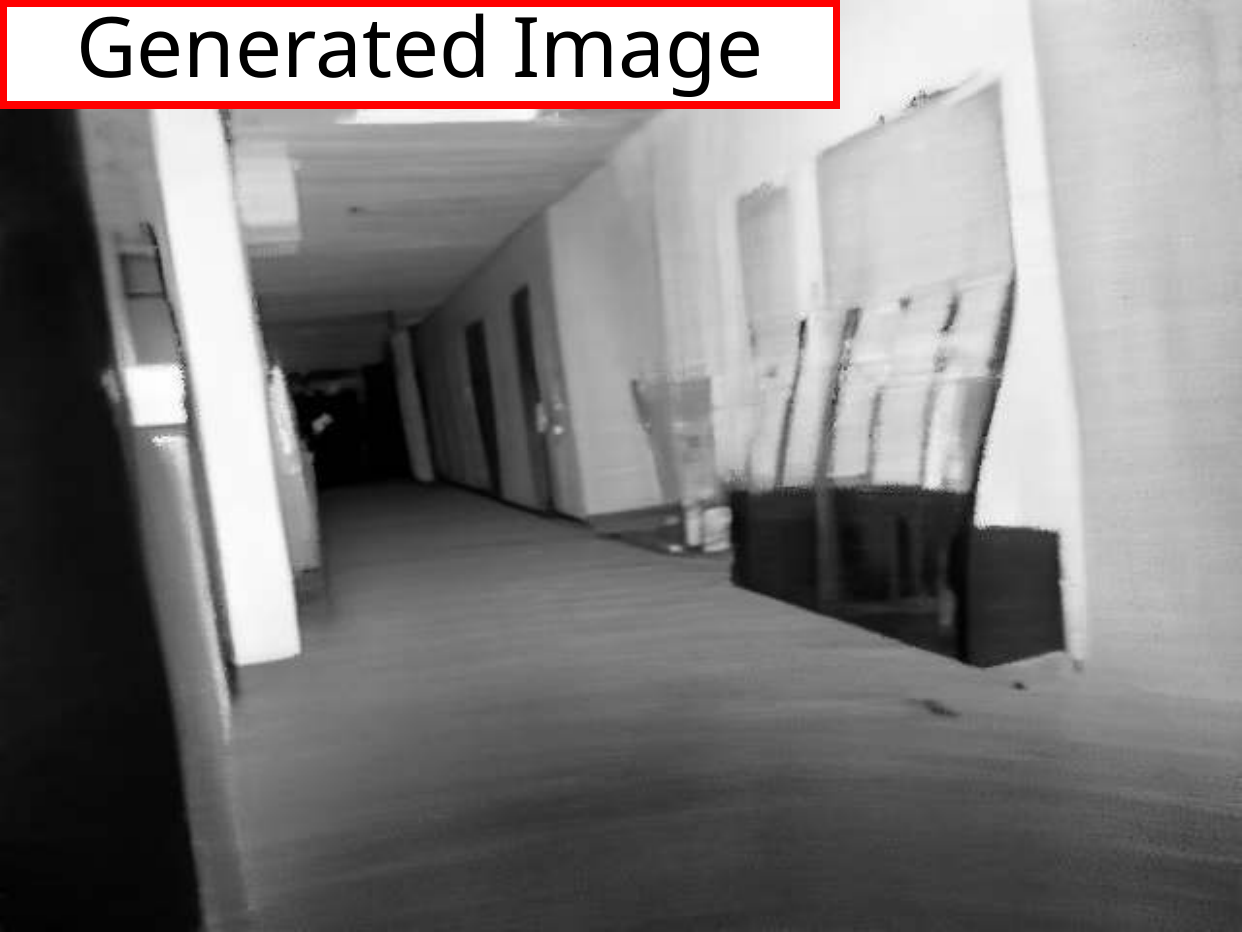} &
    \includegraphics[width=0.32\linewidth]{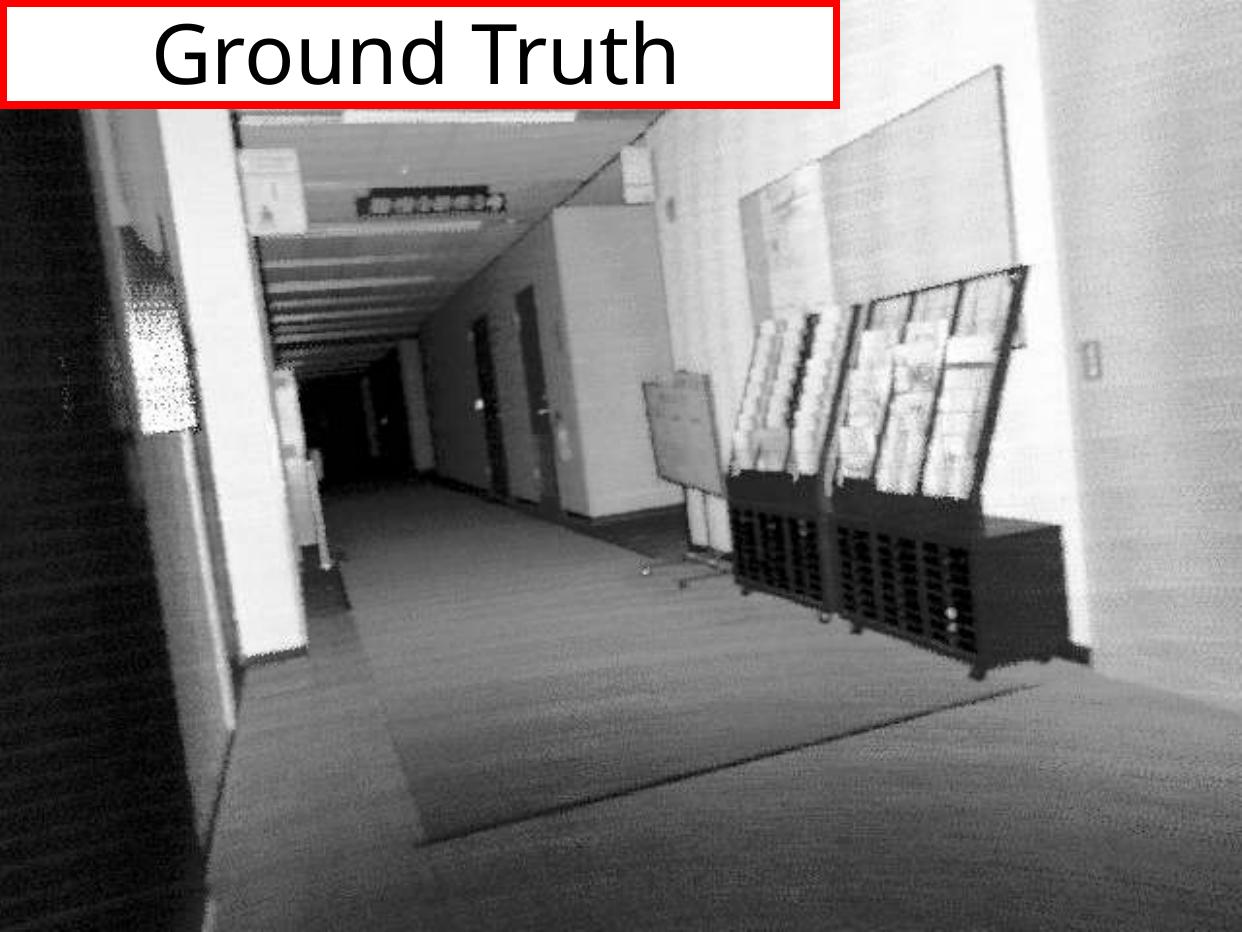} \\[-0.8ex]
    \multicolumn{3}{c}{(a) Indoor environments (Ouster OS1-64)} \\[0.8ex]
    \includegraphics[width=0.32\linewidth]{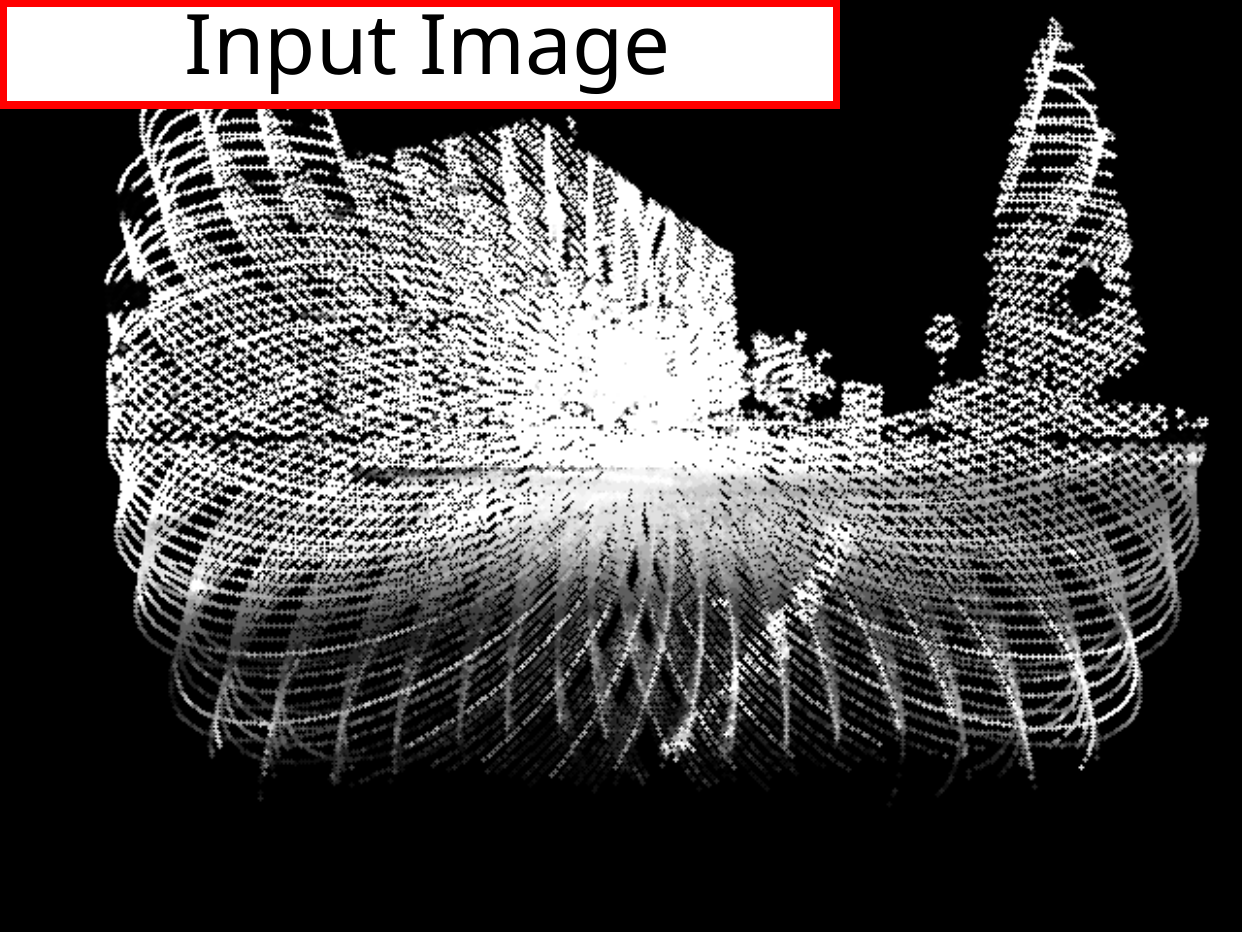} &
    \includegraphics[width=0.32\linewidth]{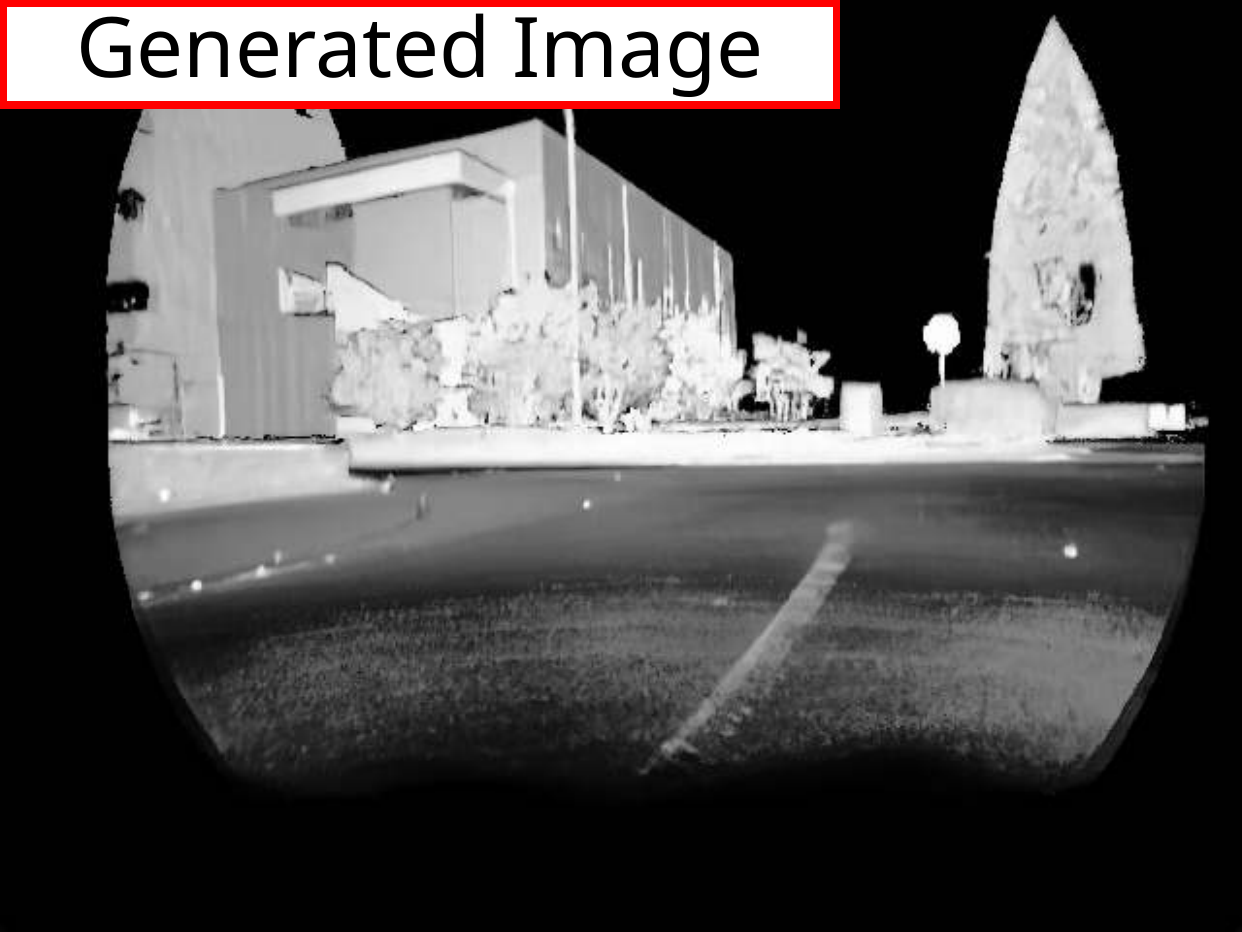} &
    \includegraphics[width=0.32\linewidth]{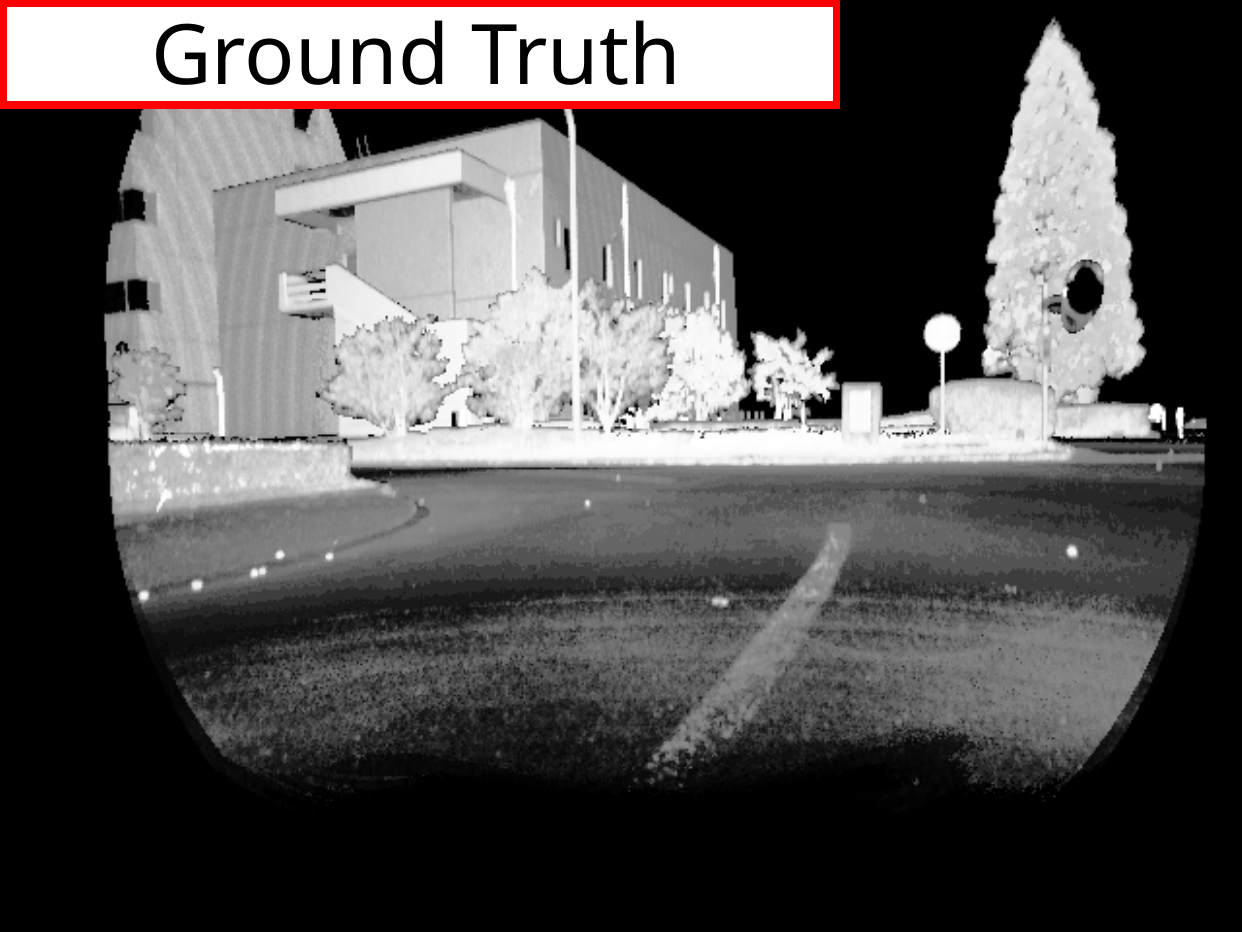} \\[-0.8ex]
    \multicolumn{3}{c}{(b) Outdoor environments (Livox AVIA)}
  \end{tabular}
  \caption{Fine-tuning results for indoor (a) and outdoor (b) environments. Columns show (left) sparse input, (middle) generated dense image, and (right) ground truth.}
  \label{fig:fine-tune}
\end{figure}

Fine-tuning our model requires dense LiDAR intensity images.
For non-repetitive LiDARs (e.g., Livox AVIA and MID360), a dense point cloud is obtained by accumulating scans while stationary, since the patterns fill the FOV over time.
For spinning LiDARs (e.g., Ouster OS1-64), however, the scan pattern is sparse and repetitive, making it difficult to obtain dense coverage from a static capture.
We therefore move the LiDAR in the vertical direction for a few seconds and accumulate points while compensating for viewpoint change and motion distortion.
We use CT-ICP\cite{dellenbach2022ct} to estimate LiDAR motion, correct distortion, and accumulate all points in the first scan's frame, yielding a dense intensity image.
We collected dense LiDAR intensity images in indoor and outdoor environments using three LiDAR sensors: Ouster OS1-64 (20 indoor and 20 outdoor locations) and Livox MID360 and AVIA (40 indoor and 40 outdoor locations each).
Fig.~\ref{fig:intensity_images} shows examples of dense LiDAR intensity images used for fine-tuning.
We apply data augmentation including image translation, rotation, horizontal flipping, and varying the virtual camera parameters used for projection.

Fig.~\ref{fig:fine-tune} shows fine-tuning results for each sensor type.
The generated images successfully capture the structure of the ground truth data.
These results demonstrate that pre-training has equipped the model with the ability to adapt to diverse scan patterns across different LiDAR sensors.

\section{Experiments}

\begin{table*}[t]
\vspace{1.45mm}
\centering
\scriptsize
\caption{Registration results on R3LIVE dataset.}
\label{registration_results}
\setlength{\tabcolsep}{1pt}
\begin{tabular}{l @{} c @{} c c @{} c @{} c c @{} c @{} c c @{} c}
\toprule
\multirow{2}{*}{Method} &
\multirow{2}{*}{Threshold ($^\circ$/m)} &
\multicolumn{3}{c}{hkust\_campus} &
\multicolumn{3}{c}{hku\_campus} &
\multicolumn{3}{c}{hku\_park} \\
\cmidrule(lr){3-5}\cmidrule(lr){6-8}\cmidrule(lr){9-11}
& & RRE ($^\circ$)$\downarrow$ & RTE (m)$\downarrow$ & RR (\%)$\uparrow$ & RRE ($^\circ$)$\downarrow$ & RTE (m)$\downarrow$ & RR (\%)$\uparrow$ & RRE ($^\circ$)$\downarrow$ & RTE (m)$\downarrow$ & RR (\%)$\uparrow$ \\
\midrule

CoFiI2P
& none/none
& $17.41 \pm 20.52$ & $9.93 \pm 26.76$ & 99.96 & $17.77 \pm 16.61$ & $5.62 \pm 12.09$ & 100.00 & $15.55 \pm 15.26$ & $3.99 \pm 10.75$ & 100.00 \\
& 45/10
& $12.16 \pm 6.51$ & $3.29 \pm 2.19$ & 81.06 & $13.53 \pm 8.47$ & $2.51 \pm 2.00$ & 84.67 & $13.30 \pm 8.74$ & $2.38 \pm 1.93$ & 93.03 \\
& 10/5
& $6.91 \pm 2.02$ & $2.15 \pm 1.11$ & 30.06 & $7.00 \pm 2.05$ & $1.72 \pm 1.00$ & 35.33 & $6.31 \pm 2.04$ & $1.58 \pm 1.00$ & 40.53 \\

\midrule

FreeReg
& none/none
& $30.32 \pm 55.55$ & $17.72 \pm 137.06$ & 99.71 & $31.02 \pm 55.89$ & $24.78 \pm 155.24$ & 99.22 & $39.38 \pm 57.80$ & $50.22 \pm 335.51$ & 99.41 \\
& 45/10
& $6.51 \pm 5.87$ & $2.26 \pm 1.77$ & 80.00 & $6.36 \pm 6.64$ & $1.50 \pm 1.30$ & 79.78 & $8.71 \pm 8.54$ & $2.08 \pm 2.01$ & 68.69 \\
& 10/5
& $4.26 \pm 2.39$ & $1.71 \pm 0.98$ & 63.13 & $3.99 \pm 2.50$ & $1.20 \pm 0.74$ & 65.89 & $3.99 \pm 2.89$ & $1.19 \pm 1.00$ & 45.73 \\

\midrule

2D3D-MATR
& none/none
& $7.92 \pm 11.69$ & $4.50 \pm 86.98$ & 99.96 & $7.04 \pm 9.13$ & $1.88 \pm 4.26$ & 100.00 & $\mathbf{8.71 \pm 14.83}$ & $\mathbf{2.07 \pm 5.50}$ & 100.00 \\
& 45/10
& $6.44 \pm 5.26$ & $1.49 \pm 1.63$ & 95.87 & $6.18 \pm 5.73$ & $1.43 \pm 1.80$ & \textbf{98.00} & $6.79 \pm 5.93$ & $1.34 \pm 1.61$ & \textbf{96.57} \\
& 10/5
& $4.41 \pm 2.37$ & $1.03 \pm 0.83$ & 76.62 & $3.98 \pm 2.49$ & $0.86 \pm 0.83$ & 79.11 & $4.56 \pm 2.30$ & $0.87 \pm 0.68$ & \textbf{77.92} \\

\midrule

Ours
& none/none
& $\mathbf{3.17 \pm 9.25}$ & $\mathbf{1.30 \pm 5.07}$ & 100.00 & $\mathbf{5.92 \pm 17.84}$ & $\mathbf{1.54 \pm 5.71}$ & 100.00 & $12.08 \pm 23.97$ & $3.23 \pm 13.08$ & 100.00 \\
& 45/10
& $\mathbf{2.30 \pm 3.54}$ & $\mathbf{0.74 \pm 1.16}$ & \textbf{97.45} & $\mathbf{2.94 \pm 4.92}$ & $\mathbf{0.73 \pm 1.07}$ & 95.56 & $\mathbf{6.50 \pm 8.49}$ & $\mathbf{1.20 \pm 1.54}$ & 91.36 \\
& 10/5
& $\mathbf{1.75 \pm 1.32}$ & $\mathbf{0.57 \pm 0.63}$ & \textbf{93.53} & $\mathbf{1.92 \pm 1.49}$ & $\mathbf{0.54 \pm 0.59}$ & \textbf{90.00} & $\mathbf{3.21 \pm 2.22}$ & $\mathbf{0.76 \pm 0.72}$ & 75.37 \\
\bottomrule
\multicolumn{11}{l}{\scriptsize \textbf{Bold} indicates the best result for each threshold setting.} \\
\end{tabular}
\end{table*}

\subsection{Experiment setup}
\subsubsection{Dataset}
R3LIVE dataset is used to evaluate registration accuracy and correspondence quality.
This dataset contains RGB images, point clouds, and IMU data recorded with a camera and a LiDAR (Livox AVIA) across various environments (e.g., walkway, park, forest).
We use hkust\_campus\_00, hku\_campus\_seq\_00, and hku\_park\_00 (hereafter hkust\_campus, hku\_campus, and hku\_park) for evaluation.
Among them, hku\_campus mainly contains structured scenes with buildings, hku\_park includes many unstructured elements such as trees and grass, and hkust\_campus contains a mixture of structured and unstructured environments.
Baseline methods requiring training were trained on hku\_main\_building, which contains a balanced variety of buildings, vegetation, and walkways.

Since the camera-LiDAR extrinsic is fixed, we create image-point cloud pairs within a 1.2\,s sliding window, producing varying displacements, ground-truth poses, and overlap per frame, which better reflects real usage.
Motion labeling and point cloud deskewing were performed using GLIM\cite{koide2024glim}, and all methods use the deskewed scans as input.

\subsubsection{Baseline Methods}
\paragraph{CoFiI2P\cite{kang2024cofii2p}}
It extracts features from images and point clouds in a hierarchical way and finds correspondences in two stages from coarse to fine.
The authors provide pre-trained weights, but because the LiDAR scan pattern in R3LIVE dataset is very different, we fine-tuned the model.

\paragraph{FreeReg\cite{wang2023freereg}}
It uses large pre-trained image-generation models to reduce the modality gap, requiring no retraining for different LiDARs or cameras.
Correspondences are built from intermediate diffusion model features, combined with monocular depth and local point cloud-based matches.

\paragraph{2D3D-MATR\cite{li20232d3d}}
It is a detector-free method that expands patch correspondences into dense matches instead of detecting separate 2D and 3D keypoints.
It learns cross-modal correlation using cross-attention for coarse matching.
We adjusted hyperparameters and retrained the model.

\subsubsection{Evaluation Metrics}
For evaluation, we measure two aspects: registration accuracy and correspondence quality.

\textbf{Registration accuracy metric}: Following CoFiI2P\cite{kang2024cofii2p}, we use Relative Rotation Error (RRE), Relative Translation Error (RTE), and Registration Recall (RR).
Let $\boldsymbol{R}^{*}$, $\boldsymbol{t}^{*}$ and $\hat{\boldsymbol{R}}$, $\hat{\boldsymbol{t}}$ denote the rotation and translation components of the ground-truth ${}^C\boldsymbol{T}_L^{*}$ and the estimated ${}^C\hat{\boldsymbol{T}}_L$, respectively.
RRE and RTE are defined as:
\begin{align}
  \text{RRE} &= \textstyle\sum_{i=1}^{3} |r(i)|, \quad \text{RTE} = \|\boldsymbol{t}^{*} - \hat{\boldsymbol{t}}\|,
\end{align}
where $\boldsymbol{r} = (r(1),r(2),r(3))$ is the Euler angle vector of $(\boldsymbol{R}^{*})^{-1}\hat{\boldsymbol{R}}$.
RR is the percentage of test pairs satisfying both $\text{RRE} < \tau_R$ and $\text{RTE} < \tau_t$.
We use the same threshold settings $(\tau_R, \tau_t)$ as in CoFiI2P\cite{kang2024cofii2p}.

\textbf{Correspondence quality metric}: Following FreeReg\cite{wang2023freereg}, we evaluate Feature Matching Recall (FMR), Inlier Ratio (IR), and Inlier Number (IN).
A 2D--3D correspondence $(\boldsymbol{v}_k, {}^L\boldsymbol{q}_k)$ is regarded as correct if the reprojection error $\|\pi({}^C\boldsymbol{T}_L^{*}\, {}^L\boldsymbol{q}_k) - \boldsymbol{v}_k\|$ is smaller than $\tau_c = 10\,\mathrm{px}$, where ${}^C\boldsymbol{T}_L^{*}$ is the ground-truth transformation.
IR is the average proportion of correct correspondences among all estimated correspondences per I2P pair.
IN is the average number of correct correspondences per I2P pair.
FMR is the fraction of I2P pairs whose IR exceeds 5\%.

\subsection{Registration Performance} \label{registration_performance}

\begin{figure*}[t]
  \vspace{1.5mm}
  \centering
  \begin{minipage}{0.24\textwidth}
    \centering
    \includegraphics[width=0.95\linewidth]{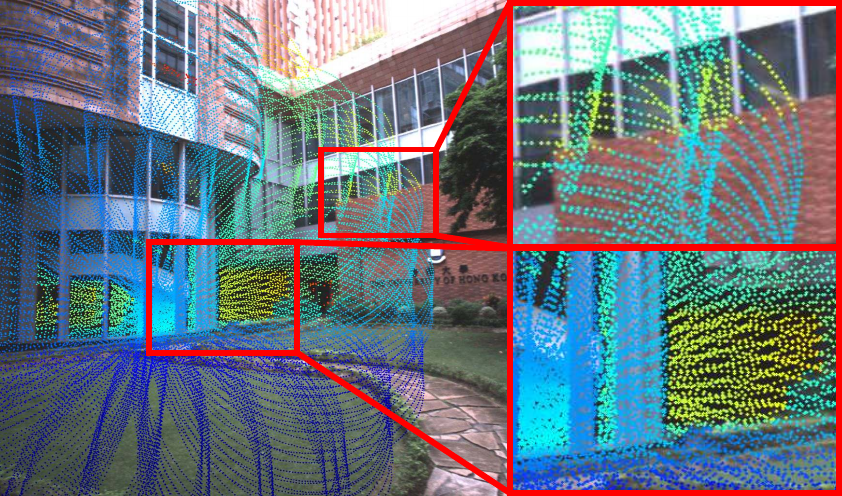}
    \par\small (a) Ground Truth
  \end{minipage}\hspace{0mm}
  \begin{minipage}{0.24\textwidth}
    \centering
    \includegraphics[width=0.95\linewidth]{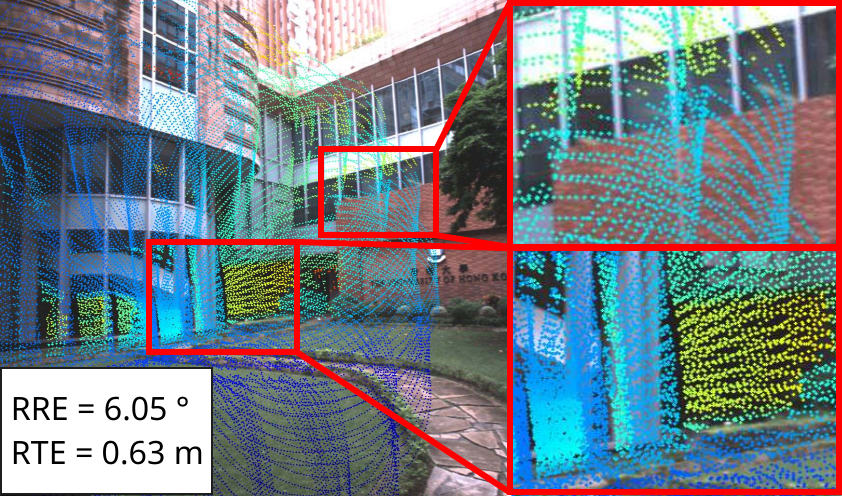}
    \par\small (b) FreeReg
  \end{minipage}\hspace{0mm}
  \begin{minipage}{0.24\textwidth}
    \centering
    \includegraphics[width=0.95\linewidth]{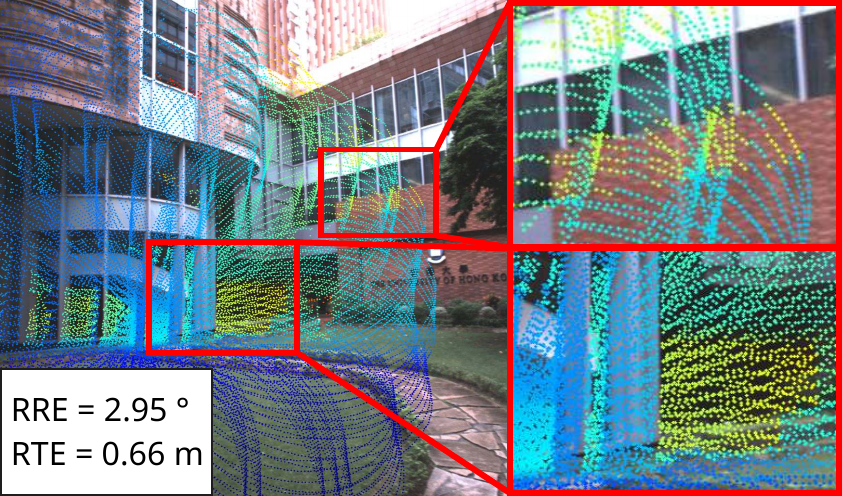}
    \par\small (c) 2D3D-MATR
  \end{minipage}\hspace{0mm}
  \begin{minipage}{0.24\textwidth}
    \centering
    \includegraphics[width=0.95\linewidth]{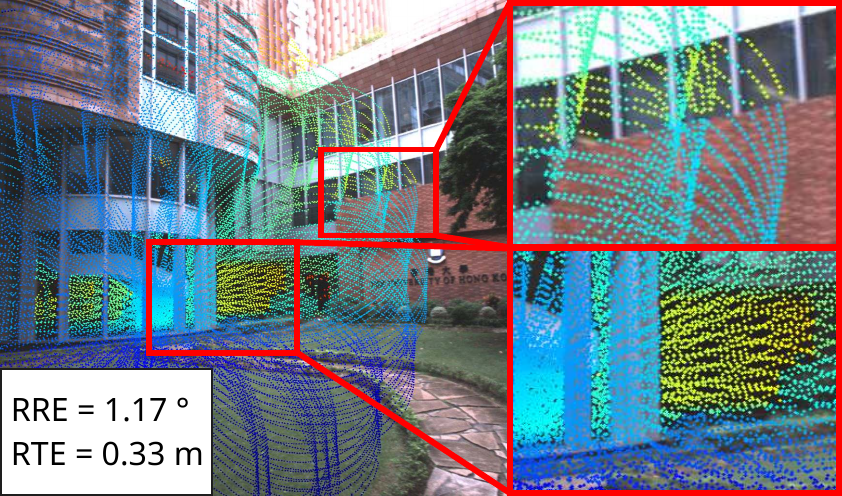}
    \par\small (d) Ours
  \end{minipage}
  \caption{Qualitative comparison of LiDAR point cloud projection onto the camera image. Points are projected using poses estimated by each method: (a) ground truth, (b) FreeReg, (c) 2D3D-MATR, and (d) our method.}
  \label{fig:registration_visualization}
\end{figure*}

\begin{figure}[t]
  \centering
  \includegraphics[width=0.80\linewidth]{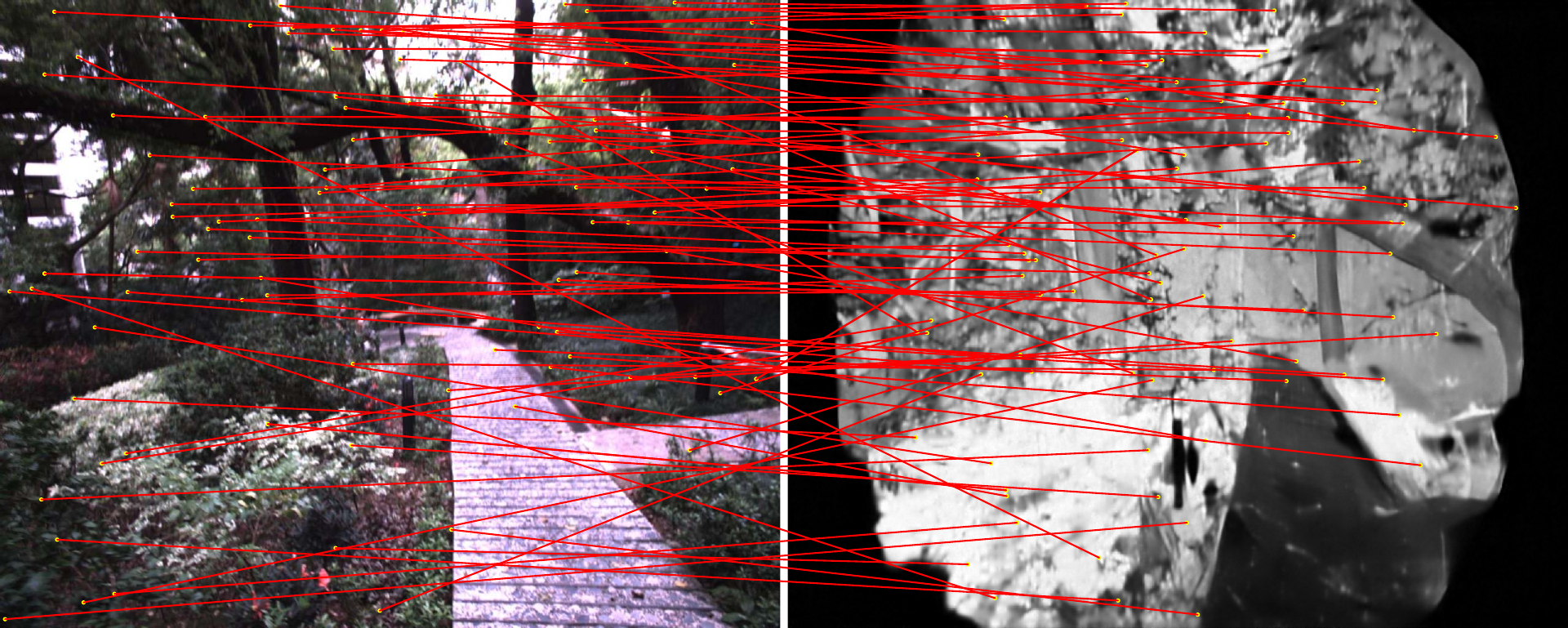}
  \caption{Failure case example in hku\_park. The left shows a camera image and the right shows the generated LiDAR intensity image. Due to dense foliage and unstable generated LiDAR intensity, numerous false feature matches are observed.}
  \label{fig:failure_case}
\end{figure}

Table~\ref{registration_results} shows registration accuracy (RRE, RTE) and registration recall (RR) for each sequence, compared against CoFiI2P, FreeReg, and 2D3D-MATR.
Our method achieves $1.75^\circ$ RRE and $0.57\,$m RTE at the $10^\circ/5\,$m threshold on hkust\_campus, outperforming all baselines.
The hku\_campus sequence also demonstrates strong alignment accuracy.
Fig.~\ref{fig:registration_visualization} further illustrates that our projected points align well with the image structure, whereas FreeReg\cite{wang2023freereg} and 2D3D-MATR\cite{li20232d3d} show noticeable misalignment.
On hku\_park, our method achieves at least 75\% RR but is slightly worse than 2D3D-MATR\cite{li20232d3d}; tree leaves in the LiDAR point clouds cause unstable intensity, leading to frequent failures (see Fig.~\ref{fig:failure_case}).
These results are achieved without training on any camera--LiDAR pairs from R3LIVE dataset, demonstrating the strong generalization of our approach, particularly in structured environments.


\subsection{Correspondence Quality}

\begin{table}[t]
\centering
\scriptsize
\caption{Correspondence quality results.}
\label{correspondence_results}
\begin{tabular}{lccc}
\toprule
Method & FMR (\%)$\uparrow$ & IR (\%)$\uparrow$ & IN (\#)$\uparrow$ \\
\midrule
CoFiI2P & 1.63 & 0.67 & 2.08 \\
FreeReg & 9.41 & 1.97 & 2.68 \\
2D3D-MATR & 30.43 & 4.10 & 25.83 \\
Ours & \textbf{48.58} & \textbf{6.80} & \textbf{41.43} \\
\bottomrule
\end{tabular}
\end{table}

\begin{figure}[t]
  \centering
  \begin{minipage}{0.49\linewidth}
    \centering
    \includegraphics[width=\linewidth]{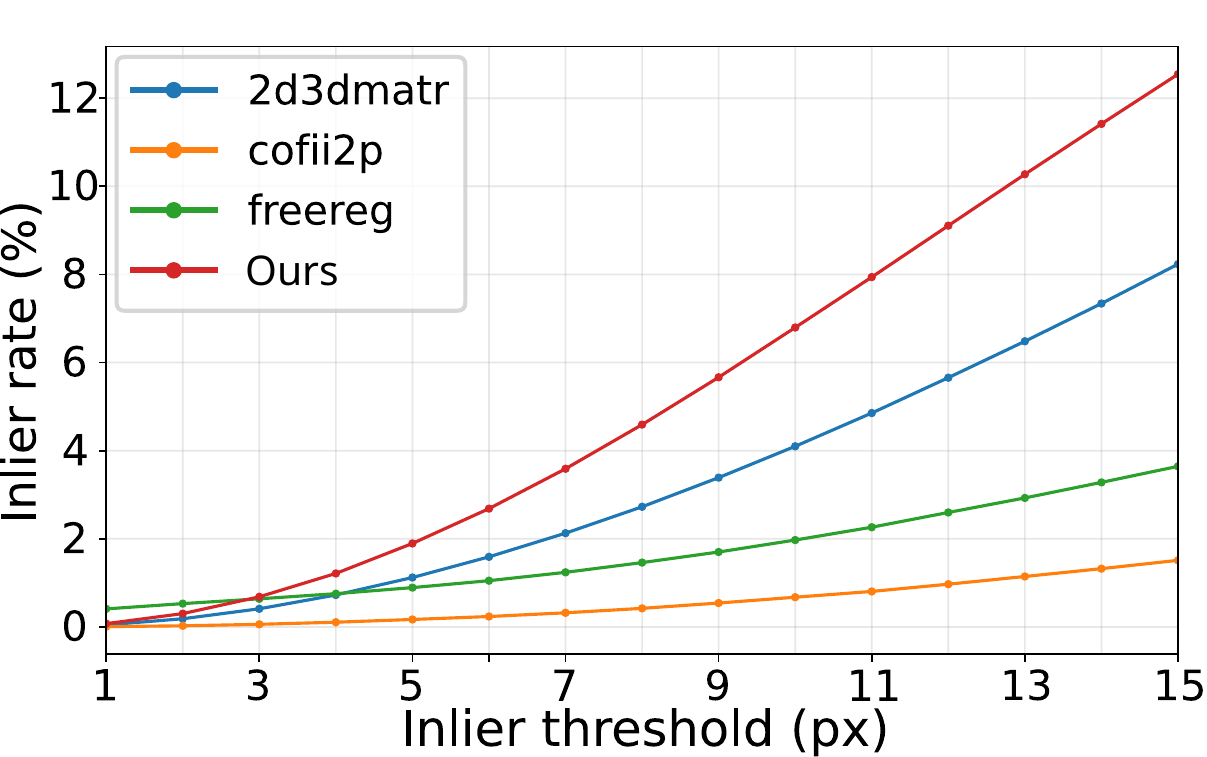}
    \par\small (a) Inlier Rate
  \end{minipage}\hspace{-1mm}
  \begin{minipage}{0.49\linewidth}
    \centering
    \includegraphics[width=\linewidth]{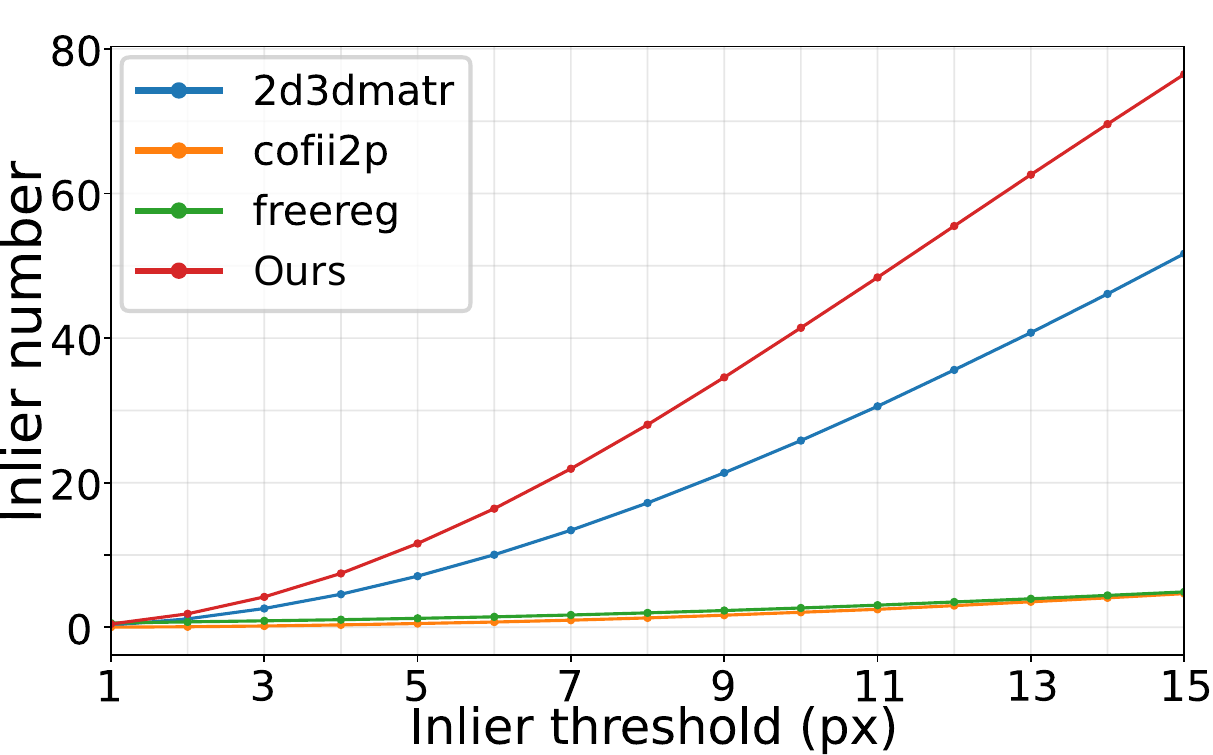}
    \par\small (b) Inlier Number
  \end{minipage}
  \caption{Inlier curves for different sequences.}
  \label{fig:inlier_curves}
\end{figure}

\begin{figure}[t]
  \centering
  \makebox[\linewidth][c]{%
    \begin{minipage}{0.48\linewidth}
      \centering
      \includegraphics[width=\linewidth]{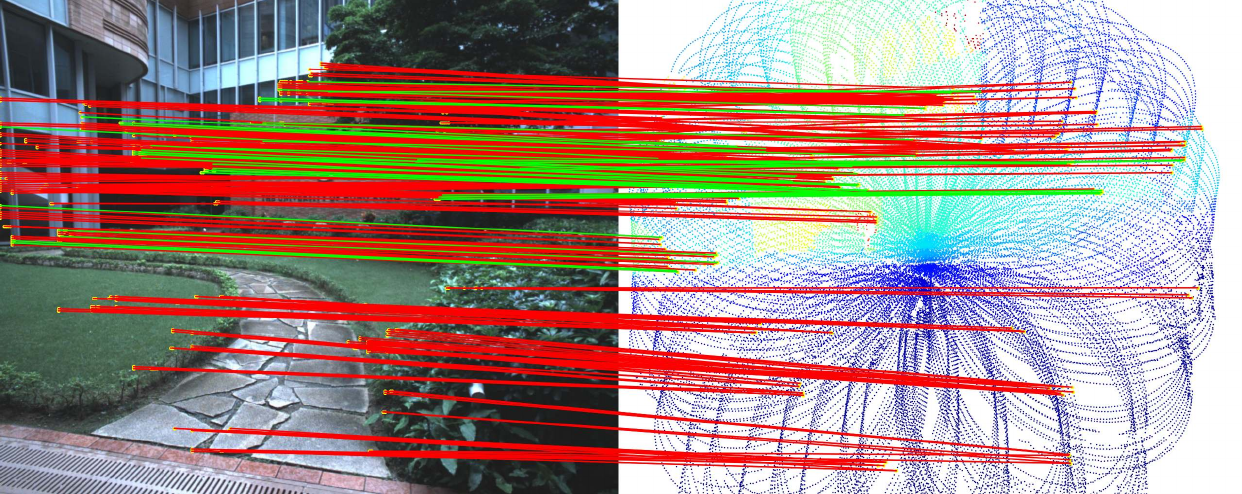}
      \par\small (a) 2D3D-MATR
    \end{minipage}\hspace{1mm}%
    \begin{minipage}{0.48\linewidth}
      \centering
      \includegraphics[width=\linewidth]{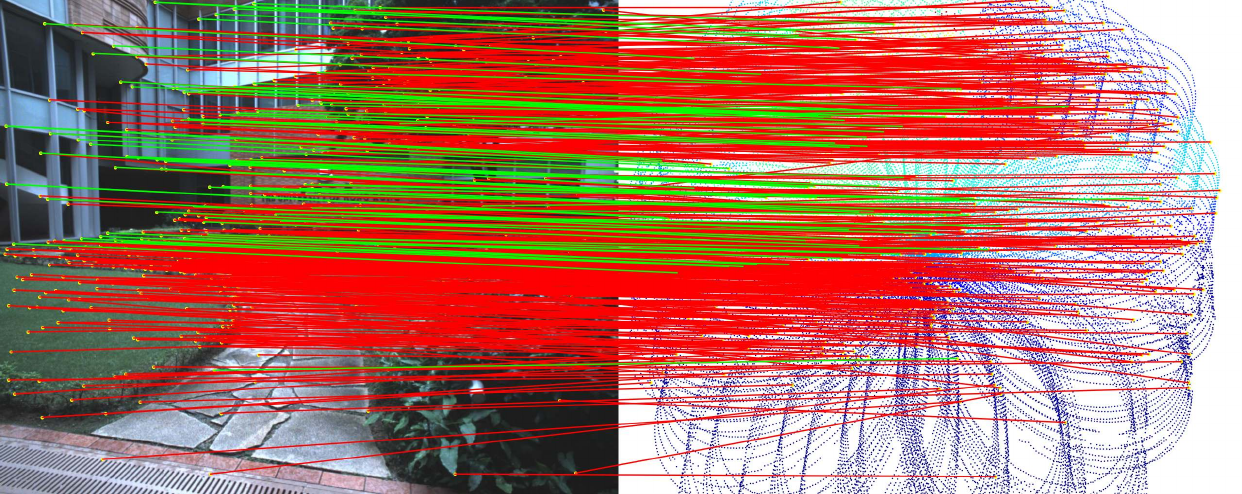}
      \par\small (b) Ours
    \end{minipage}%
  }
  \caption{Qualitative comparison of feature correspondences on hku\_campus. (a) 2D3D-MATR and (b) our method.}
  \label{fig:correspondence_comparison}
\end{figure}

Table~\ref{correspondence_results} reports correspondence quality (FMR, IR, IN) as an intermediate evaluation of the generated intensity image quality and feature matches.
Our method achieves the best results across all metrics; its IN is $1.6\times$ that of 2D3D-MATR and its FMR approaches 50\%, confirming that the upsampled intensity images provide sufficient texture for cross-modal feature matching.
The inlier curves in Fig.~\ref{fig:inlier_curves} further confirm this: CoFiI2P and FreeReg yield almost no inliers at any threshold, whereas our method obtains more than 10 inliers at a 5\,px threshold.
Fig.~\ref{fig:correspondence_comparison} qualitatively shows that our method produces richer correspondences than 2D3D-MATR.

\subsection{Computation Cost}

\begin{table}[t]
  \centering
  \scriptsize
  \caption{Computation cost comparison.}
  \label{tab:computation_cost}
  \begin{tabular}{lcc}
    \toprule
    Method & Iteration time (s) & VRAM (MB) \\
    \midrule
    CoFiI2P   & \textbf{0.28}          & \textbf{1,991} \\
    FreeReg   & 22.62         & 10,209 \\
    2D3D-MATR & 0.70          & 1,976 \\
    Ours      & 0.68 & 3,252 \\
    \bottomrule
  \end{tabular}
\end{table}

Table~\ref{tab:computation_cost} reports per-frame computation time and VRAM usage evaluated on an NVIDIA RTX A5000 24GB.
Our method runs at 0.68\,s per frame, faster than 2D3D-MATR (0.70\,s) and FreeReg (22.62\,s).
This speed is sufficient for use as a lower-rate matching module alongside high-frequency LiDAR odometry and visual odometry.
For VRAM usage, our method requires 3,252\,MB, much lower than FreeReg (10,209\,MB), which also employs a generative model.

\subsection{Ablation Study}

\subsubsection{Generation Steps}

\begin{table}[t]
  \centering
  \scriptsize
  \caption{Ablation study on different generation steps.}
  \label{tab:generation_steps}
  \begin{tabular}{lccccc}
    \toprule
    Generation Steps & Threshold ($^\circ$/m) & RRE ($^\circ$)$\downarrow$ & RTE (m)$\downarrow$ & RR (\%)$\uparrow$ & Time (s)$\downarrow$ \\
    \midrule
    1 step          & none/none & 4.89           & 1.63           & 100.00          & \\
                    & 45/10     & \textbf{2.99}  & 0.80           & 96.28           & \textbf{0.68} \\
                    & 10/5      & \textbf{1.96}  & \textbf{0.59}  & 90.31           & \\
    \midrule
    5 steps         & none/none & \textbf{4.67}  & 1.39           & 100.00          & \\
                    & 45/10     & 3.22           & 0.80           & 97.25           & 1.24 \\
                    & 10/5      & 2.02           & 0.62           & 90.78           & \\
    \midrule
    10 steps        & none/none & 4.79           & \textbf{1.23}  & 100.00          & \\
                    & 45/10     & 3.21           & \textbf{0.79}  & \textbf{97.31}  & 1.86 \\
                    & 10/5      & 2.07           & 0.63           & \textbf{90.84}  & \\
    \bottomrule
  \end{tabular}
\end{table}

\begin{figure}[t]
  \vspace{1.5mm}
  \centering
  \begin{minipage}{0.28\linewidth}
    \centering
    \includegraphics[width=\linewidth]{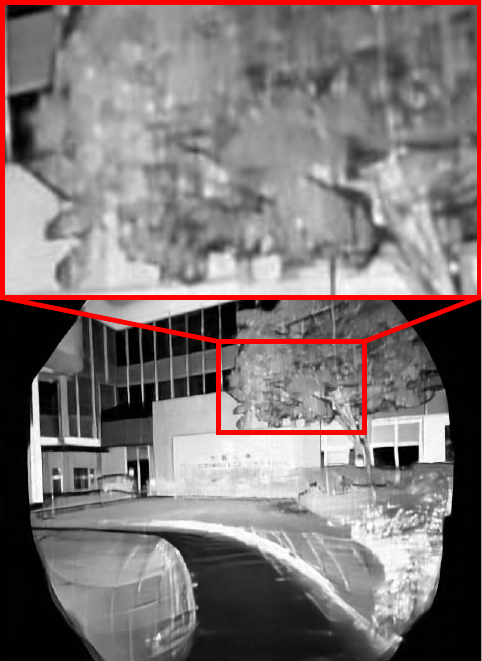}
    \par\small (a) 1 step
  \end{minipage}\hspace{0mm}
  \begin{minipage}{0.28\linewidth}
    \centering
    \includegraphics[width=\linewidth]{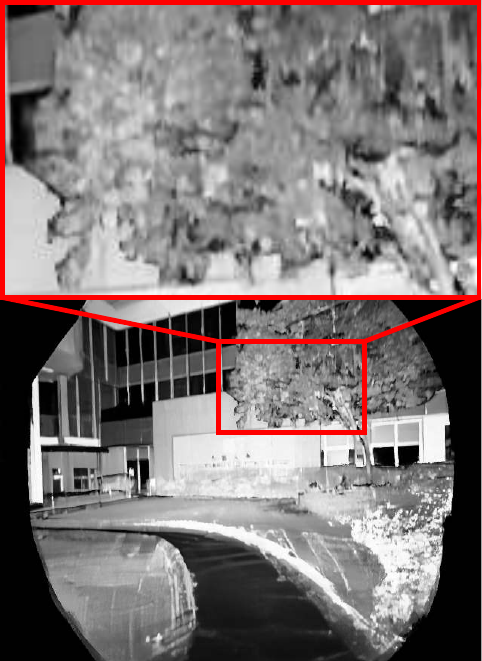}
    \par\small (b) 5 steps
  \end{minipage}\hspace{0mm}
  \begin{minipage}{0.28\linewidth}
    \centering
    \includegraphics[width=\linewidth]{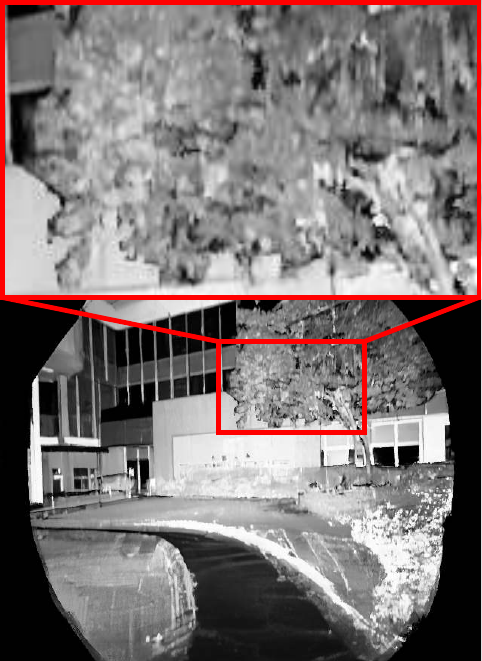}
    \par\small (c) 10 steps
  \end{minipage}
  \caption{Generated dense LiDAR intensity images with (a) 1, (b) 5, and (c) 10 generation steps.}
  \label{fig:generation_step_comparison}
  \vspace{2mm}
\end{figure}

To evaluate the effect of generation steps, we ran experiments with 1, 5, and 10 steps.
The number of steps causes only minor changes in registration performance, and a single step already achieves sufficient accuracy.
Fig.~\ref{fig:generation_step_comparison} shows the upsampled intensity images for each setting.
The 1-step result is blurrier than the 5- or 10-step results, but overall structures such as building walls and the ground are clearly represented, so it contains enough information for feature matching in pose estimation.
There is little difference between the 5- and 10-step outputs, and 5 steps are sufficient to fully realize the model's image generation capability.
Increasing the number of generation steps tends to slightly worsen RRE and RTE within the $10^\circ/5\,$m threshold, possibly because the generative model produces overly sharpened edges with more steps, increasing false correspondences.

\subsubsection{Pre-Training}

\begin{table}[t]
  \centering
  \scriptsize
  \caption{Ablation study on the effect of pre-training.}
  \label{tab:three_methods_rre_rte_ir}
  \begin{tabular}{lccc}
    \toprule
    Method & RRE ($^\circ$)$\downarrow$ & RTE (m)$\downarrow$ & IR (\%)$\uparrow$ \\
    \midrule
    2D3D-MATR                 & 7.92          & 3.78          & 4.10 \\
    Ours                      & \textbf{4.89} & \textbf{1.63} & \textbf{6.80} \\
    Ours (w/o pre-training)   & 18.65         & 10.64         & 1.24 \\
    \bottomrule
  \end{tabular}
\end{table}

To evaluate the effect of pre-training, we tested a model without pre-training.
Table~\ref{tab:three_methods_rre_rte_ir} shows the comparison results.
The model without pre-training showed a large performance drop (RRE $18.65^\circ$, RTE $10.64\,$m, IR $1.24\%$), demonstrating the importance of pre-training.
Since the fine-tuning data were collected only in-house and are very limited (only 80 scenes), the non-pre-trained model likely overfitted the fine-tuning set, leading to poor generalization.
\subsubsection{Matching}

\begin{table}[t]
  \vspace{1.5mm}
  \centering
  \scriptsize
  \caption{Effect of depth search radius (DSR).}
  \label{tab:depth_search_radius_rre_rte_ir}
  \begin{tabular}{lccc}
    \toprule
    DSR (pixels) & RRE ($^\circ$)$\downarrow$ & RTE (m)$\downarrow$ & IR (\%)$\uparrow$ \\
    \midrule
    0 & 8.72          & 2.29          & \textbf{6.98} \\
    1 & 5.16          & \textbf{1.57} & 6.79 \\
    3 & \textbf{4.89} & 1.63 & 6.80 \\
    5 & 4.90          & 1.63 & 6.79 \\
    \bottomrule
  \end{tabular}
\end{table}

Since our method upsamples only intensity, points without depth must be approximated by nearby depth values; we evaluate the effect of the search radius for this lookup (Table~\ref{tab:depth_search_radius_rre_rte_ir}).
We tested radii of 0, 1, 3, and 5 pixels.
With radius 0 (discarding matches without depth), accuracy dropped to RRE $8.72^\circ$ and RTE $2.29\,$m.
With radius 5 accuracy $4.90^\circ$ was worse than with radius 3. 
This likely happens because a larger search radius increases depth approximation error and degrades pose estimation. 

\subsection{Different Sensors}

\begin{figure}[t]
  \vspace{1.0mm}
  \centering
  \begin{minipage}{0.29\linewidth}
    \centering
    \includegraphics[width=\linewidth]{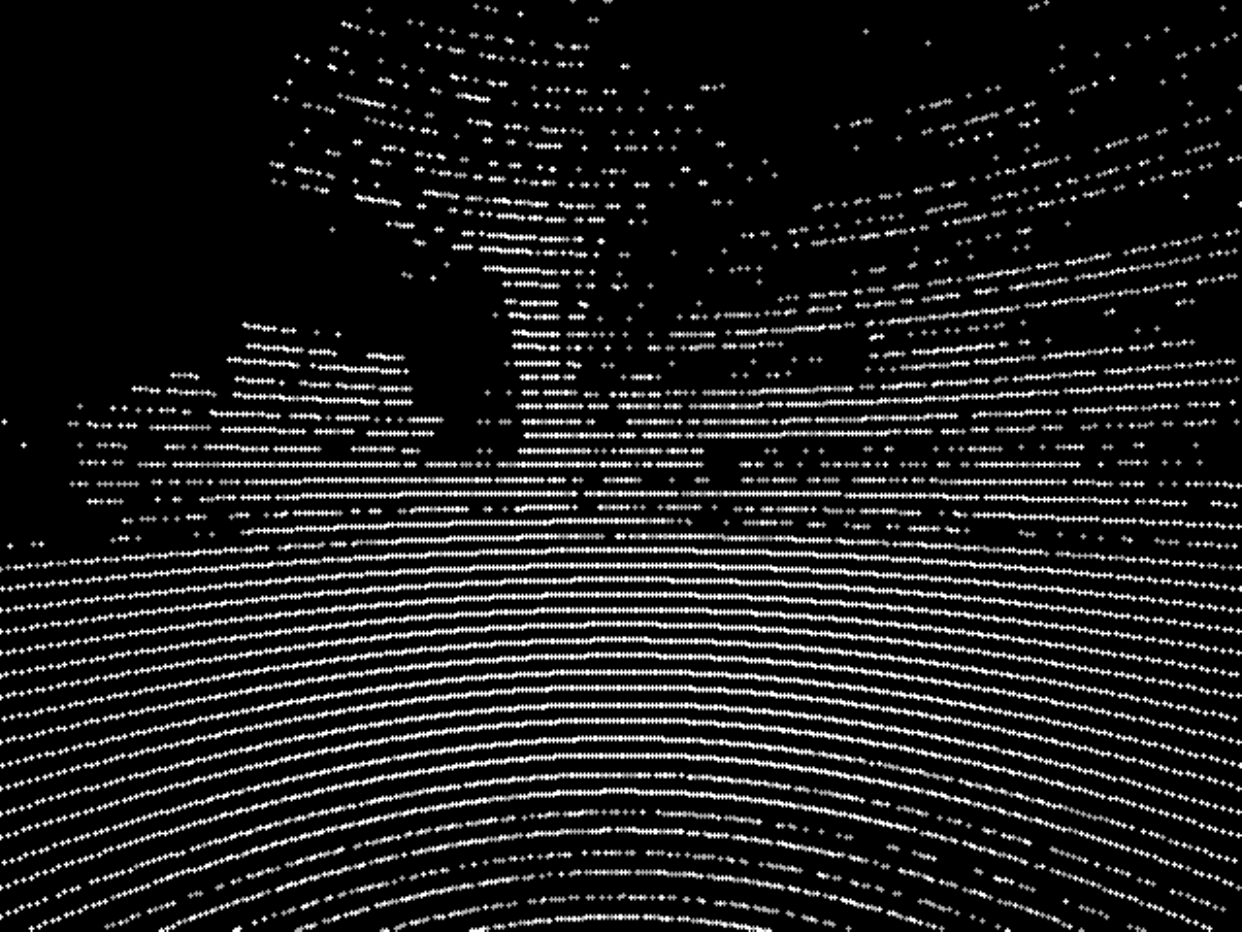}
    \par\small (a) Sparse
  \end{minipage}\hspace{0mm}
  \begin{minipage}{0.29\linewidth}
    \centering
    \includegraphics[width=\linewidth]{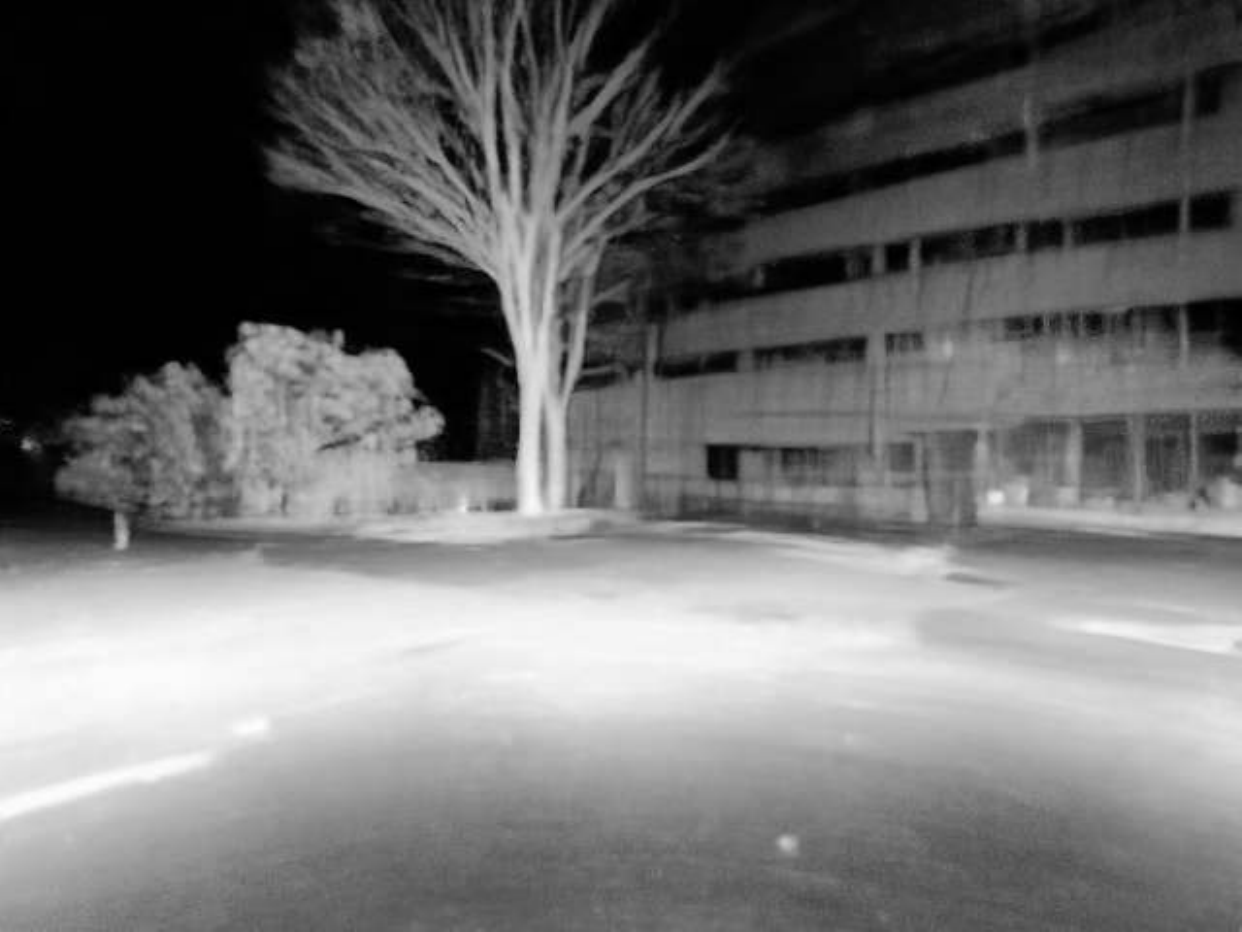}
    \par\small (b) Upsampled
  \end{minipage}\hspace{0mm}
  \begin{minipage}{0.29\linewidth}
    \centering
    \includegraphics[width=\linewidth]{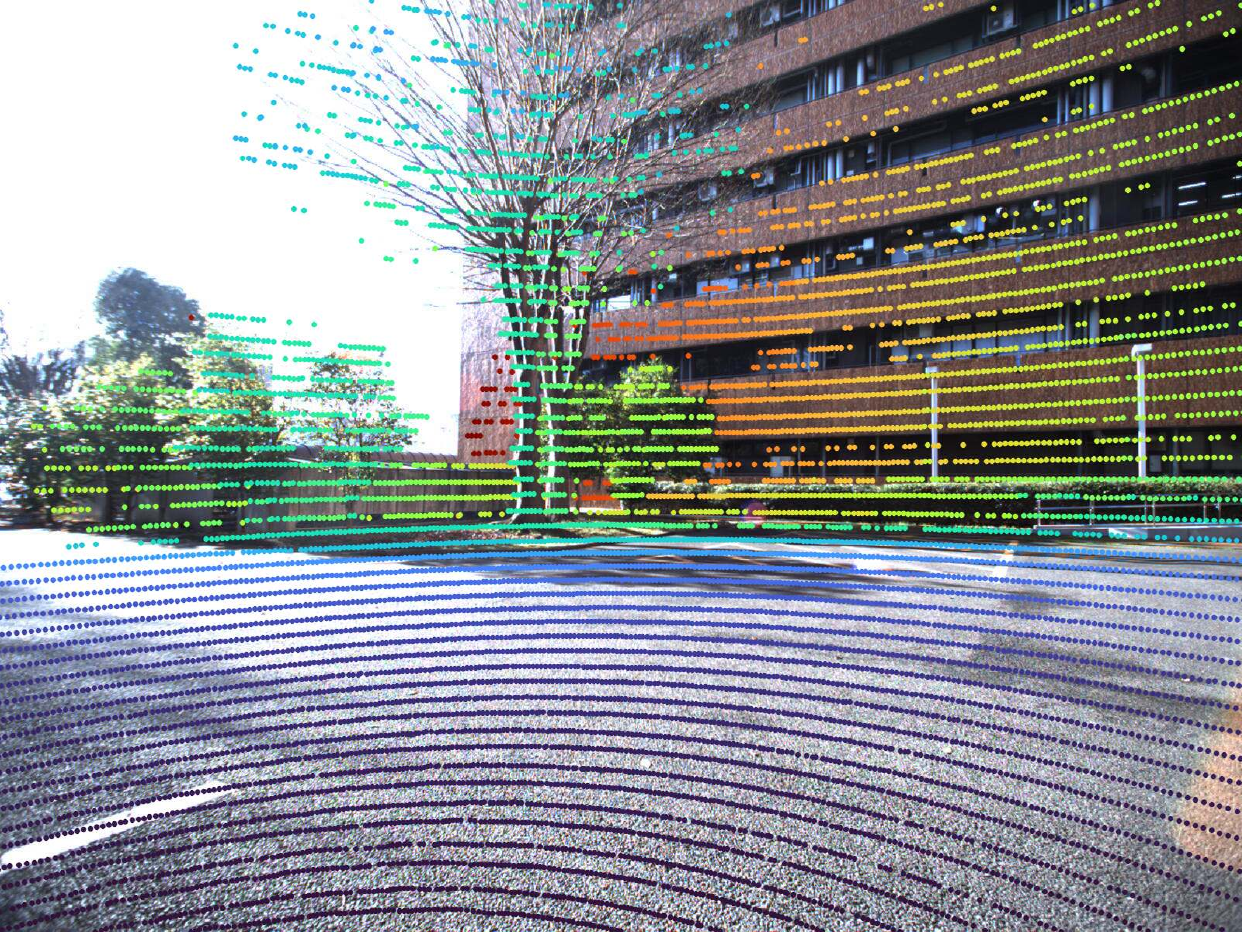}
    \par\small (c) Result
  \end{minipage}
  \caption{Examples for Ouster OS1-64. (a) Sparse intensity image, (b) upsampled intensity image, and (c) projected points using the estimated pose.}
  \label{fig:os1_examples}
\end{figure}

To demonstrate that the method generalizes beyond Livox AVIA used in the R3LIVE dataset evaluation, we show qualitative results on the Ouster OS1-64, a spinning LiDAR, in Fig.~\ref{fig:os1_examples}.
The sparse intensity images from a single scan are line-sparse, but we observe they are densely inpainted.
From the projected images using estimated poses, we can see the method adapts well to spinning-type LiDAR.
Although we used a pinhole camera in this work, the method can be extended to omnidirectional cameras.

\section{Conclusion}
We presented an I2P registration method that treats LiDAR as an imaging sensor: sparse LiDAR scans are upsampled to dense intensity images via Conditional Rectified Flow and matched to camera images using pre-trained feature matchers.
Experiments on the R3LIVE dataset demonstrated high accuracy and strong generalization.
Limitations include degraded performance in unstructured environments.
Future work includes integration with more advanced matchers and correspondence-free registration models.



\section*{Acknowledgment}
This paper is based on results obtained from the BRIDGE Program (R7-H05), implemented by the Cabinet Office, Government of Japan.

\bibliographystyle{IEEEtran}
\bibliography{refs}

\end{document}